\documentclass{article}
\usepackage[utf8]{inputenc}
\usepackage{url}
\usepackage{graphicx}
\usepackage{subfigure}
\usepackage{caption}
\usepackage{a4wide}
\usepackage{latexsym}
\usepackage{colortbl}
\usepackage{multirow}
\usepackage{rotating}
\usepackage{amssymb}
\usepackage{amsmath}
\usepackage{booktabs}
\DeclareMathOperator{\TF}{\textsf{tf}}
\DeclareMathOperator{\IDF}{\textsf{idf}}

\begin{document}

\title{A Case Study of Spanish Text Transformations for Twitter Sentiment Analysis}
\author{Eric S. Tellez$^{1,3}$ \and Sabino Miranda-Jiménez$^{1,3}$ \and Mario Graff$^{1,3}$ \and Daniela Moctezuma$^{1,2}$ \and Oscar S. Siodia$^2$ \and Elio A. Villase\~nor$^1$}
\date{%
$^1$INFOTEC Centro de Investigaci\'on e Innovaci\'on en Tecnolog\'ias
de la Informaci\'on y Comunicaci\'on, Circuito Tecnopolo Sur No 112, Fracc. Tecnopolo Pocitos II, Aguascalientes 20313, M\'exico\\
$^2$CentroGEO Centro de Investigaci\'on en Ciencias de Informaci\'on Geoespacial,
Circuito Tecnopolo Norte No. 117, Col. Tecnopolo Pocitos II, C.P., Aguascalientes, Ags 20313 M\'exico\\
$^3$CONACyT Consejo Nacional de Ciencia y Tecnolog\'ia,
Direcci\'on de Cátedras, Insurgentes Sur 1582, Cr\'edito Constructor, Ciudad de M\'exico 03940 M\'exico~\\ ~\\
This work is published in Expert Systems with Applications 
\url{https://doi.org/10.1016/j.eswa.2017.03.071}}

\maketitle

\begin{abstract}

Sentiment analysis is a text mining task that determines the polarity of a given text, i.e., its positiveness or negativeness. Recently, it has received a lot of attention given the interest in opinion mining in micro-blogging platforms. These new forms of textual expressions present new challenges to analyze text given the use of slang, orthographic and grammatical errors, among others. Along with these challenges, a practical sentiment classifier should be able to handle efficiently large workloads.

The aim of this research is to identify which text transformations (lemmatization, stemming, entity removal, among others), tokenizers (e.g., words $n$-grams), and tokens weighting schemes impact the most the accuracy of a classifier (Support Vector Machine) trained on two Spanish corpus. The methodology used is to exhaustively analyze all the combinations of the text transformations and their respective parameters to find out which characteristics the best performing classifiers have in common. Furthermore, among the different text transformations studied, we introduce a novel approach based on the combination of word based $n$-grams and character based $q$-grams. The results show that this novel combination of words and characters produces a classifier that outperforms the traditional word based combination by $11.17\%$ and $5.62\%$ on the INEGI and TASS'15 dataset, respectively.

\end{abstract}

\section{Introduction}
\label{sec:introduction}

In recent years, the production of textual documents in social media has increased exponentially; for instance, up to April 2016, Twitter has 320 million active users, and Facebook has 1,590 million users.\footnote{http://www.statista.com/statistics/272014/global-social-networks-ranked-by-number-of-users/} In social media, people share comments about many disparate topics, i.e., events, persons, and organizations, among others. These facts have had the result of seeing social media as a gold mine of human opinions, and consequently, there is an increased interest in doing research and business activities around opinion mining and sentiment analysis fields.

Automatic sentiment analysis of texts is one of the most important tasks in text mining, where the goal is to determine whether a particular document has either a positive, negative or neutral opinion\footnote{Albeit, there are other variations considering intermediate levels for sentiments, e.g. more {\em positive} or less {\em positive}}. Determining whether a text document has a positive or negative opinion is becoming an essential tool for both public and private companies, \cite{BLiuBook2015,Peng2008}. Given that it is a useful tool to know {\em what people think} about anything; so, it represents a major support for decision-making processes (for any level of government, marketing, etc.)~\cite{PangLee2008}.

Sentiment analysis has been traditionally tackled as a classification task where two major problems need to be faced. Firstly, one needs to transform the text into a suitable representation, this is known as text modeling. Secondly, one needs to decide which classification algorithm to use; one of the most widely used is Support Vector Machines (SVM). This contribution focus on the former problem, i.e., we are interested in improving the classification by finding a suitable text representation.

Specifically, the contribution of this research is twofold. Firstly, we parametrize our text transformations with different techniques such as: lemmatization, stemming, and entity removal, just to mention a few (Table~\ref{tab/parameters} contains all the transformations explored). This parametrization is used to exhaustively evaluate the entire configurations space to know those transformations that produce the best SVM classifier on two sentiment analysis corpus written in Spanish. Counterintuitively, we found that the complexity of techniques used in the pre-processing step is not correlated with the final performance of the classifier, e.g., a classifier using lemmatization, which is one of the pre-processing techniques having the greatest complexity, might not be one of the systems having the highest performance.

Secondly, we propose a novel approach based on the combination of word based $n$-grams and character based $q$-grams. This novel combination of words and characters produces a classifier that outperforms the traditional word based combination by $11.17\%$ and $5.62\%$ on the INEGI and TASS'15 dataset, respectively. Hereafter, we will use {\em n-words} to refer to word $n$-grams, and {\em $q$-grams} to character $q$-grams just to make a clear distinction between these techniques.

This manuscript is organized as follows. Section~\ref{sec:introduction} introduces the paper and the problem being tackled. Section~\ref{sec:Related_Work} deals with literature review. The text transformations are described in Section \ref{sec:preprocessing}, meanwhile the parameters settings and definition of the problem are presented on Section \ref{sec:problem_parameters}. Section~\ref{sec:results} describes our experimental results. Finally, Section~\ref{sec:discussion} and  Section~\ref{sec:conclusions} present the discussion and conclusions of our results along with possible directions for future work.

\section{Related Work}
\label{sec:Related_Work}

The sentiment analysis task has been widely studying due to the interest to know the people's opinions and feelings about something, particularly in social media. This task is commonly tackled in two different ways. The first one involves the use of static resources that summarize the sense or semantic of the task; these knowledge databases contain mostly affective lexicons. These lexicons are created by experts, in psychology or by automated processes, that perform the selection of features (words) along with a corpus of labeled text as done in \cite{Ghiassi2013}. Consequently, the task is solved by trying to detect how the affective features are used in a text, and how these features can be used to predict the polarity of a given text.

The alternative approach states the task as a text classification problem. This includes several distinguished parts like the pre-processing step, the selection of the vectorization and weighting schemes, and also the classifier algorithm. So, the problem consists of selecting the correct techniques in each step to create the sentiment classifier. Under this approach, the idea is to process the text in a way that the classifier can take advantage of the features to solve the problem. Our contribution focus in this later approach; we describe the best way to pre-process, tokenize, and vectorize the text, based on a fixed set of text-transformation functions. For simplicity, we fix our classifier to be Support Vector Machines (SVM). SVM is a classifier that excels in high dimensional datasets as is the case of text classification, \cite{joachims1998text}. This section reviews the related literature.


There are several works in the sentiment analysis literature which use several representations; such as dictionaries~\cite{Alam2016206},~\cite{khan2016sentimi}; text content and social relations between
users~\cite{wu2016structured}; relationships between meanings of a word in a corpus~\cite{Hossein2014}; co-occurrence patterns of
words~\cite{Saif20165}, among others.

Focusing on the $n$-grams technique, a method that considers the local context of the word sequence and the semantic of the whole sentence is proposed in \cite{cui2015sentiment}.
The local context is generated via the ``bag-of-n-words'' method, and the sentence's sentiment is determined based on the individual contribution of each word. The word embedding is learned from a large monolingual corpus through a deep neural network, and the n-words features are obtained from the word embedding in combination with a sliding window procedure.

A hybrid approach that uses n-gram analysis for feature extraction together with a dynamic artificial neural network for sentiment analysis is proposed in \cite{Ghiassi2013}. Here, a dataset over $10,000,000$ of tweets, related to Justin Bieber topic, was used.
As a result, a Twitter-specific lexicon with a reduced feature set was obtained.

The work presented in \cite{Han2013CodeXCA} proposes an approach for sentiment analysis which combines an SVM classifier and a wide range of features like bag-of-word (1-words, 2-words) and part-of-speech (POS) features, etc., as well as votes derived from
character n-words language models to achieve the final result.
The authors concluded that lexical features (1-words, 2-words) produce the better contributions.


In \cite{Abinash2016} different classifiers and representations were applied to determine the sentiment in movie reviews, taken from internet blogs.
The classifiers tested were Naive Bayes, maximum entropy, stochastic gradient descent, and SVM.
These algorithms use n-words, for $n$ in $\{1,2,3\}$ and all the combinations.
Here, the results show that the value of $n$ increases the classification accuracy decreases, i.e., using 1-words and 2-words the result achieved is better than using 3-words, 4-words, and 5-words.

Regarding the use of $q$-grams; in \cite{Aisopos2011} a method that captures textual patterns is introduced. This method creates a graph, whose nodes correspond to $q$-grams of a document and their edges denoted the average distance between them. A comparative analysis on data from Twitter is performed between three representation models: term vector model, $q$-grams, and $q$-grams graphs.
The authors found that vector models are faster, but $q$-grams (especially 4-grams) perform better in terms of classification quality.

With the purpose to attend sentiment analysis in Spanish tweets, a number of works has been presented in the literature, e.g. several sizes of n-grams and some polarity lexicons combined with a Support Vector Machine (SVM) was used in \cite{almeidadeustotech}. Another approach which uses polarity lexicons with a number of features related to n-words, part-of-speech tag, hashtags, emoticon and lexicon resources is described in \cite{araque2015aspect}.

Features related to lexicons and syntactic structures are commonly used, for example, \cite{alvarez2015gti}, \cite{camara2015sinai}, \cite{deensemble}, \cite{bordabittenpotato},\cite{deas2015spanish}.
In the other hand, features related to word vectorization, e.g. Word2Vec and Doc2Vec, are also used in several works, such as \cite{diaz2015participacion,valverde2015comparing}.

Following with the Spanish language, in the most recent TASS (Taller de An\'alisis de Sentimientos '16) competition, was presented some works still using polarity dictionaries and vectorization approach; such is the case of \cite{casasola2016evaluacion}, where an adaptation of Turney dictionary \cite{Turney2002} over 5 millions of Spanish tweets was generated. Furthermore, \cite{casasola2016evaluacion} in the step of vectorization uses n-grams and skip-grams in combination with this polarity dictionary.
\cite{quiros2016labda} proposes the use of word embedding with SVM classifier. Despite the explosion of words using word embeddings, the classical word vectorization is still in use, cite{montejo2016participacion}.

A new approach is using ensembles or a combination of several techniques and classifiers, e.g. the work presented in \cite{ceron2016jacerong} proposes an ensemble built on the combination of systems with the lowest correlation between them.
\cite{Hurtado2016} presents another ensemble method where the  Tweetmotif's tokenizer, \cite{tweetmotif}, is used in conjunction with Freeling \cite{padro2012freeling}. These tools create a vector space that is the input for an SVM classifier.


It can be seen that one of the objectives of the related work is to optimize the number of n-words or $q$-grams (almost tackled as independent approaches), to increase performance; clearly, there is not a consensus. This lack of agreement motivates us to perform an extensive experimental analysis of the effect of the parameters (including $n$ and $q$ values), and so, we determined the best parameters on the Twitter databases employed.

\section{Text Representation}
\label{sec:preprocessing}

\begin{figure*}
		\centering
		\includegraphics[width=0.9\textwidth]{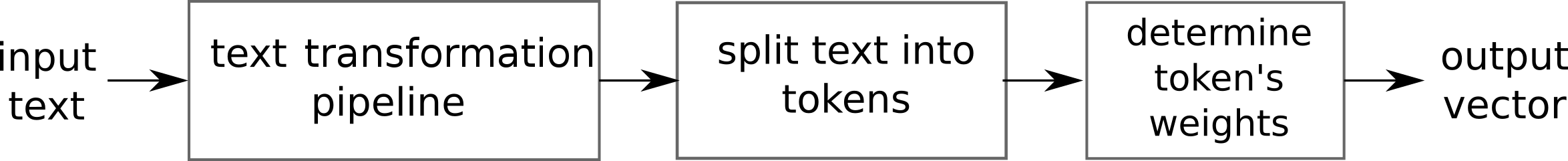}
		\caption{Generic treatment of input text to obtain the input vectors for the classifier.}
		\label{fig:general_procedure}
	\end{figure*}

Natural Language Processing (NLP) is a broad and complex area of knowledge having many ways to represent an input text~\cite{Giannakopoulos2012,encyclopediaML}. In this research, we select the widely used vector representation of a text given its simplicity and powerful representation. Figure~\ref{fig:general_procedure} depicts the procedure used to transform a text input into a vector. There are three main blocks: the first one transforms the text into another text representation, then the text is tokenized, and, finally, the vector is calculated using a weighting scheme. The resulting vectors are the input of the classifier.

In the following subsections, we described the text transformation techniques used which have a counterpart in many languages, the proper implementation of them rely heavily on the targeted language, in our case study the Spanish language. The interested reader looking for solutions in a particular language is encouraged to follow the relevant linguistic literature for its objective language, in addition to the general literature in NLP~\cite{Jurafsky2009,NLTK2009,encyclopediaML}.

	\subsection{Text Transformation Pipeline}
	\label{sec:textTrans}

One of the contributions of this manuscript is to measure the effects that each different text transformation has on the performance of a classifier. This subsection describes the text transformations explored whereas the particular parameters of these transformations can be seen in Table~\ref{tab/parameters}.

\subsubsection{\textsf{TFIDF} (\textsf{tfidf})}

In the vector representation, each word, in the collection, is associated with a coordinate in a high dimensional space. The numeric value of each coordinate is sometimes called the {\em weight} of the word. Here, $\TF\times\IDF$ (Term Frequency-Inverse Document Frequency) \cite{BYRNmir1999} is used as bootstrapping weighting procedure. More precisely, let $D = \{D_1, D_2, \dots, D_N\}$ be the set of all documents in the corpus, and $f^i_w$ be the frequency of the word $w$ in document $D_i$. $\TF^i_w$ is defined as the normalized frequency of $w$ in $D_i$ \[\TF^i_w = \frac{f^i_w}{\max_{u \in D_i}\{f^i_u\}}. \] In some way, $\TF$ describes the importance of $w$, locally in $D_i$. On the other hand, $\IDF$ gives a global measure of the importance of $w$; \[\IDF_w = \log \frac{N}{\left|\{D_i \mid f^i_w > 0\} \right|}.\]

The final product, $\TF \times \IDF$, tries to find a balance between the local and the global importance of a term. It is common to use variants of $\TF$ and $\IDF$ instead of the original ones, depending in the application domain~\cite{encyclopediaML}. Let $v_i$ be the vector of $D_i$, a weighted matrix \textsf{TFIDF} of the collection $D$ is created by concatenating all individual vectors, in some consistent order. Using this representation, a number of machine learning methods can be applied; however, the plain transformation of text to \textsf{TFIDF} poses some problems.	On one hand, all documents will contain common terms having a small semantic content such as articles and determiners, among others. These terms are known as {\em stopwords}. The bad effects of stopwords are controlled by \textsf{TFIDF}, but most of them can be directly removed since they are fixed for a given language. On the other hand, after removing stopwords, \textsf{TFIDF} will produce a very high dimensional vector space, $O(N)$ in Twitter, since new terms are commonly introduced (e.g. misspellings, URLs, hashtags). This will rapidly yield to the {\em Curse of Dimensionality}, which makes hard to learn from examples since any two random vectors will be orthogonal with high probability. From a more practical point of view, a high dimensional representation will also impose huge memory requirements, at the point of being impossible to train a typical implementation of a machine learning algorithm (not being designed to use sparse vectors).


\subsubsection{Stopwords (\textit{del-sw})}

In many languages, like Spanish, there is a set of extremely common words such as determiners or conjunctions ($the$ or $and$) which help to build sentences but do not carry any meaning themselves. These words are known as \textit{Stopwords}, and they are removed from the text before any attempt to classify them. A stop list is built using the most frequent terms from a huge document collection. We used the Spanish stop list included in NLTK Python package~\cite{NLTK2009}.

\subsubsection{Spelling}

Twitter messages are full of slang, misspelling, typographical and grammatical errors among others; however, in this study, we focus only on the following transformations:

\begin{description}
    \item[Punctuation (\textit{del-punc}).] This parameter considers the use of symbols such as question mark, period, exclamation point, commas, among other spelling marks.
    \item[Diacritic (\textit{del-diac}).] The Spanish language is full of diacritic symbols, and its wrong usage is one of the main sources of orthographic errors in informal texts. Thus, this parameter considers the use or absence of diacritical marks.
    \item[Symbol reduction (\textit{del-d1, del-d2}).] Usually, twitter messages use repeated characters to emphasize parts of the word to attract user's attention. This aspect makes the vocabulary explodes. Thus, we applied two strategies to deal with these phenomena: the first one replaces the repeated symbols by one occurrence of the symbol, and the second one replaces the repeated symbols by two occurrences to keep the word emphasize at the minimal level.
    \item[Case sensitive (\textit{lc}).] This parameter considers letters to be normalized in lowercase or to keep the original text. The aim is to cut the words that are the same in uppercase and lowercase.
\end{description}

\subsubsection{Stemming (\textit{stem})}

Stemming is a heuristic process in Information Retrieval field that chops off the end of words and often includes the removal of derivational affixes. This technique uses the morphology of the language coded in a set of rules; to find out word stems and reduce the vocabulary collapsing derivationally related words. In our study, we use the Snowball Stemmer for the Spanish language implemented in NLTK package~\cite{NLTK2009}.

\subsubsection{Lemmatization (\textit{lem})}

Lemmatization process is a complex task from  Natural Language Processing that determines the lemma of a group of word forms, i.e., the dictionary form of a word. For example, the words \textit{went} and \textit{goes} are the verb conjugations of the verb \textit{go}; and these words are grouped under the same lemma \textit{go}. To apply this process, we use Freeling tool~\cite{freeling} as Spanish lemmatizer. All texts are prepared by the \textit{Error correction} process before applying lemmatization to obtain the best results of part-of-speech identification.

\paragraph{Error correction} Freeling is a tool for text analysis, but the assumption is that text is well-written. However, language used in Twitter is very informal, with slang, misspellings, new words, creative spelling, URLs, specific abbreviations, hashtags (which are especial words for tagging in Twitter messages), and emoticons (which are short strings and symbols that express different emotions).
These problems are treated to prepare and standardize tweets for the lemmatization stage to get the best results. All words in each tweet are checked to be a valid Spanish word or are reduced according to the rules for Spanish word formation.

In general, words or tokens with invalid duplicated vowels or consonants are reduced to valid or standard Spanish words, e.g., (\em{ruiiidoooo} $\rightarrow$ \em{ruido} (noise); \em{jajajaaa} $\rightarrow$ \em{jaja}; \em{jijijji} $\rightarrow$ \em{jaja}). We used an approach based on Spanish dictionary, a statistical model for common double letters, and heuristic rules for common interjections. In general, the duplicated vowels or consonants are removed from the target word; the resulting word is looked up in a Spanish dictionary (approximately 550,000 entries) to be validated, it is included in Freeling. For words that are not in the dictionary are reduced at least with valid rules for Spanish word formation. Also, colloquial words and abbreviations are transformed using a regular expression based on a dictionary of those sort of words, figure~\ref{fig/example-colloquial} illustrates some rules. The text on the left side of the arrow is replaced by the text of the right side.     Twitter tags such as user names, hashtags (topics), URLs, and emoticons are handled as special tags in our representation to keep the structure of the sentence.

\begin{figure*}[h!]
\centering
\fbox{
    \begin{minipage}{0.63\textwidth}
    \scriptsize
    \hspace{0.28cm} {\em tqm} $\rightarrow$ {\em te quiero mucho} (I love you so much),\\
    {\em compu} $\rightarrow$ {\em computadora} (computer).
    \end{minipage}
}
\caption{Expansion of colloquial words and abbreviations.}
\label{fig/example-colloquial}
\end{figure*}

\begin{figure*}
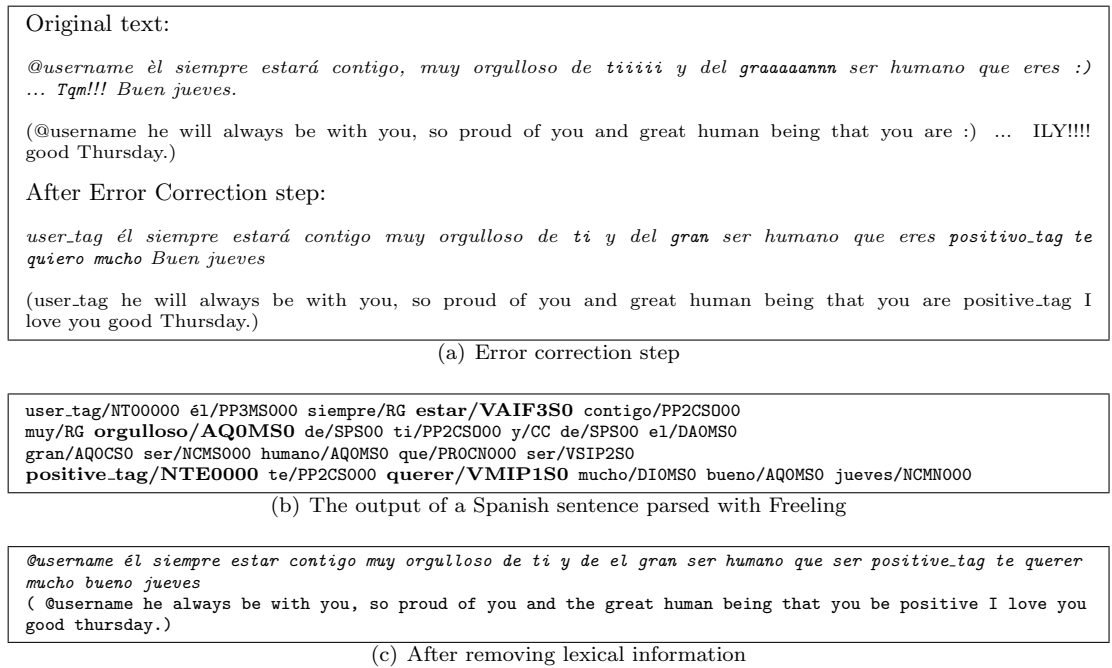

\subfigure[Error correction step] {
    \fbox{
        \begin{minipage}{0.95\textwidth}
        \small
        Original text:\\~\\
        \scriptsize
        {\em @username \`el siempre estar\'a contigo,  muy orgulloso de \texttt {tiiiii} y del \texttt{graaaaannn} ser humano que eres \texttt {:)} ... \texttt {Tqm}!!! Buen jueves.}
        \\ ~\\
        (@username he will always be with you, so proud of you and great human being that you are :) ... ILY!!!! good Thursday.)
        \\ ~\\
        \small
        After Error Correction step:\\~\\
        \scriptsize
        {\em user\_tag \'el siempre estar\'a contigo muy orgulloso de \texttt {ti} y del \texttt {gran} ser humano que eres \texttt {positivo\_tag te quiero mucho} Buen jueves}
        \\ ~\\
        (user\_tag he will always be with you, so proud of you and great human being that you are positive\_tag I love you good Thursday.)

        \end{minipage}
    }
    \label{fig/error-example}
}
\subfigure[The output of a Spanish sentence parsed with Freeling] {
    \fbox{
        \begin{minipage}{0.95\textwidth}
        \scriptsize
        {\tt
        user\_tag/NT00000 \'el/PP3MS000 siempre/RG {\bf estar/VAIF3S0} contigo/PP2CSO00 \\
        muy/RG {\bf orgulloso/AQ0MS0} de/SPS00 ti/PP2CSO00 y/CC de/SPS00 el/DA0MS0 \\
        gran/AQ0CS0 ser/NCMS000 humano/AQ0MS0 que/PR0CN000 ser/VSIP2S0  \\
        {\bf positive\_tag/NTE0000} te/PP2CS000 {\bf querer/VMIP1S0} mucho/DI0MS0 bueno/AQ0MS0 jueves/NCMN000
        }
        \end{minipage}
    }
    \label{fig/pos-example}
}
\subfigure[After removing lexical information]{
    \fbox{
        \begin{minipage}{0.95\textwidth}
        \small
         \scriptsize
         {\tt
        {\em @username \'el siempre \textbf{estar} contigo muy orgulloso de ti y de  el gran ser humano que \textbf{ser} positive\_tag te \textbf{querer} mucho \textbf{bueno} jueves} \\
        ( @username he always be with you, so proud of you and the great human being that you be positive I love you good thursday.)
        }
        \end{minipage}
    }
    \label{fig/examples-filtering}
}

\caption{A step-by-step lemmatization of a tweet.}
\label{fig/example-lemma}
\end{figure*}

In Figure~\ref{fig/example-lemma}, we can see the lemmatized text after applying Freeling. As we mentioned, the text is prepared with the Error correction step (see Figure~\ref{fig/error-example}) then Freeling is applied to normalize words. Figure~\ref{fig/pos-example} shows Freeling's output where each token has the original word followed by the slash symbol and its lexical information. The lexical information can be read as follows; for instance, token {\tt orgulloso/AQ0MS0} (proud) stands for  adjective as part of speech (AQ), masculine gender  (M), and singular number (S); the token {\tt querer/VMIP1S0} (to want) stands for lemmatized main verb as part of speech (VM), indicative mood (I), present time (P), singular	form of the first person (1S); {\tt positive\_tag/NTE0000} stands for noun tag as part of speech, and so on.

Lexical information is used to identify entities, stopwords, content words among others, it depends on the settings of the other parameters. The words are filtered based on heuristic rules that take into account the lexical information shown in Fig.~\ref{fig/pos-example}. Finally, lexical information is removed in order to get the lemmatized text depicted on Figure~\ref{fig/examples-filtering}.

\subsubsection{Negation (\textit{neg})}

Spanish negation markers might change the polarity of the message. Thus, we attached the negation clue to the nearest word, similar to the approaches used in \cite{sidorov2012}. A set of rules was designed for common Spanish negation structures that involve negation markers, namely, {\em no} (not), {\em nunca, jam\'as} (never), and {\em sin} (without). The rules are processed in order, and, when one of them matches, the remaining rules are discarded. We have two sorts of rules; it depends on the input text. If the text is not parsed by Freeling, a few rules (regular expressions) are applied to negate the nearest word to the negation marker using only the information on the text, e.g., avoiding pronouns and articles. The second approach uses a set of fine-grained rules to take advantage of the lexical information, approximately 50 rules were designed considering the negation markers. The negation marker is attached to the closest word to the marker.

In the box below, Pattern 1 and Pattern 2 are examples of negation rules (regular expressions). A rule consists of two parts: the left side of the arrow represents the text to be matched, and the right side of the arrow is the structure to be replaced. All rules are based on a linguistic motivation taking into account lexical information. The set of negation rules are available\footnote{\url{http://ws.ingeotec.mx/~sadit/}}.

\begin{figure*}[t]
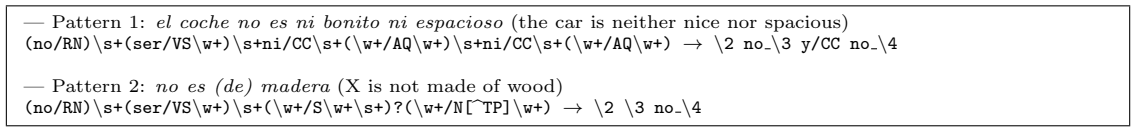

	~\\{\noindent\fbox{
			\begin{minipage}{0.97\textwidth}
				\scriptsize
				--- Pattern 1: {\em el coche no es ni bonito ni espacioso} (the car is neither nice nor spacious)\\
				{\tt (no/RN)\textbackslash s+(ser/VS\textbackslash w+)\textbackslash s+ni/CC\textbackslash s+(\textbackslash w+/AQ\textbackslash w+)\textbackslash s+ni/CC\textbackslash s+(\textbackslash w+/AQ\textbackslash w+) $\rightarrow$ \textbackslash 2 no\_\textbackslash 3 y/CC no\_\textbackslash 4
				}\\~\\
				--- Pattern 2: {\em no es (de) madera} (X is not made of wood)\\
				\scriptsize
				{\tt (no/RN)\textbackslash s+(ser/VS\textbackslash w+)\textbackslash s+(\textbackslash w+/S\textbackslash w+\textbackslash s+)?(\textbackslash w+/N[$\widehat{~}$TP]\textbackslash w+) $\rightarrow$ \textbackslash 2 \textbackslash 3 no\_\textbackslash 4
				}
			\end{minipage}
		}~\\}

\caption{An example of negation rules}
\label{fig/negation-rules}
\end{figure*}

For example, in the sentence {\em El coche no es ni bonito ni espacioso} (The car is neither nice nor spacious), the negation marker {\em no} is attached to its two adjectives {\em no\_bonito} (not nice) and {\em no\_espacioso} (not spacious), as it is showed in Pattern 1, the negation marker is attached to group 3 (\textbackslash3) and group 4 (\textbackslash4) that stand for adjective position because of the coordinating conjunction {\em ni}. The number of group is identified by parenthesis in the rule from left to right. Negation markers are attached to content words (nouns, verbs, adjectives, and some adverbs), e.g., {\em `no seguir'} ({\em	do not follow}) is replaced by {\em `no\_seguir'}, {\em `no es bueno'} ({\em it is not good}) is replaced by {\em `es no\_bueno'}, {\em `sin comida'} ({\em without food}) is replaced by {\em`no\_comida'}. Figure~\ref{fig/negation-rules} exemplifies a pair of these negation rules.

\subsubsection{Emoticon (\textit{emo})}

In the case of emotions, we classify more than 500 popular emoticons, including text emoticons, and the whole set of emoticons (close to 1600) defined by~\cite{EmotUnicode2016} into three classes: positive, negative or neutral, which are replaced by a polarity word or definition associated to the emoticon according to the Unicode standard. The emoticons considered as positive are replaced by the word \textit{positive}, negative emoticons are replaced by the word \textit{negative}, neutral emotions are replaced by the word \textit{neutral}. Emoticons that do not have a polarity, or are ambiguous, are replaced by the associated Unicode text.
    Table~\ref{table/emoticons} shows an excerpt of the dictionary that maps emoticons to their corresponding polarity class.

	\begin{table}
		\centering
		\begin{tabular}{ccc}
			{\tt :) ~~ :D ~~  :P} 	 & ~ $\rightarrow$    &	\textsf{positive} \\
			{\tt :( ~~ :-( ~~ :'(}   & ~ $\rightarrow$ &	\textsf{negative} \\
			{\tt :-| ~~ U$\_$U ~~ -.-} & ~ $\rightarrow$     &	\textsf{neutral} \\
			\textsf {emoticon without polarity} & ~ $\rightarrow$ & {\em unicode-text}\\ 
		\end{tabular}~\\~\\
		\caption{An excerpt of the mapping table from Emoticons to its polarity words. }
		\label{table/emoticons}
	\end{table}
\subsubsection{Entity (\textit{del-ent})}

We consider entities to be proper names, hashtags, urls or nicknames. However, nicknames (see \textit{usr} parameter, Table~\ref{tab/parameters}) is a particular feature in Twitter messages; thus, user names is another parameter to see the effect on the classification system. User names, urls and numbers (see \textit{url, num}) parameters, Table~\ref{tab/parameters}) could be grouped under an especial generic name. Entities such as user names and hashtags are identified directly by its corresponding especial symbol $@$ and $\#$, and proper names are identified using Freeling, the lexical information used to identify a proper name is ``NP0000".

\subsubsection{Word-based n-grams (\textit{n-words})}

N-words are widely used in many NLP tasks, and they have also been used in sentiment analysis by~\cite{sidorov2012, cui2015sentiment}. N-words are word sequences. To compute the n-words, the text is tokenized and n-word are calculated from tokens. NLTK Tokenizer is used to identified word tokens. For example, let $T=\texttt{"the lights and shadows of your future"}$, its 1-words (unigrams) are each word alone, and its 2-words (bigrams) set are the sequences of two words, the set ($W^T_2$), and so on. For example, let $W^T_2 = \{ \texttt{\small the lights, lights and, and shadows, shadows of, of your, your future}\},$ then, given a text of $m$ words, we obtain a set with at most $m-n+1$ elements. Generally, n-words are used up to 2 or 3-words because it is uncommon to find good matches of word sequences greater than three or four words~\cite{Jurafsky2009}.

\subsubsection{Character-based $q$-grams (\textit{q-grams})}

In addition to the traditionally n-words representation, we represent the resulting text as $q$-grams. A $q$-grams is an agnostic language transformation that consists in representing a document by all its substring of length $q$. For example, let $T=\texttt{abra\_cadabra}$, its 3-grams set are	\[Q^T_3 = \{ \texttt{abr, bra, ra\_, a\_c, \_ca, aca, cad, ada, dab}\},\] 	so, given text of size $m$ characters, we obtain a set with at most $m-q+1$ elements. Notice that this transformation handle white-spaces as part of the text. Since there will be $q$-grams connecting words, in some sense, applying $q$-grams to the entire text can capture part of the syntactic information in the sentence. The rationale of $q$-grams is to tackle misspelled sentences from the approximate pattern matching perspective~\cite{NRbook02}, where it is used for efficient searching of text with some degree of error.

A more elaborated example shows why the q-gram transformation is more robust to variations of the text. Let
$T=\texttt{I\_like\_vanilla}$ and $T' = \texttt{I\_lik3\_vanila}$, clearly, both texts are different and a plain
algorithm will simply associate a low similarity between both texts. However, after extracting its 3-grams, the resulting
objects are more similar:\\
$Q^T_3 = \{ \texttt{I\_l, \_li, lik, ike, ke\_, e\_v, \_va, van, ani, nil, ill, lla} \}$\\
$Q^{T'}_3 = \{ \texttt{I\_l, \_li, lik, ik3, k3\_, 3\_v, \_va, van, ani, nil, ila} \}$

Just to fix ideas, let these two sets to be compared using the Jaccard's coefficient as similarity, i.e.
$$\frac{|Q^T_3 \cap Q^{T'}_3|}{|Q^T_3 \cup Q^{T'}_3|} = 0.448.$$ These sets are more similar than the ones resulting from
the original text split as words $$\frac{|\{\texttt{I, like, vanilla}\} \cap \{\texttt{I, lik3, vanila}\}|}{|\{\texttt{I, like, vanilla}\} \cup \{\texttt{I, lik3, vanila}\}|} = 0.2$$
The assumption is that a machine learning algorithm knowing how to classify $T$ will do a better job classifying $T'$ using $q$-grams than a plain representation.
This fact is used to create a robust method against misspelled words and other deliberated modifications to the text.

\subsection{Examples of Text Transformation Stage}
	\label{sec:Example-textTrans}

In order to illustrate the text transformation pipeline, we show the examples in Figure \ref{fig/example1} and Figure \ref{fig/example2}.
In Figure~\ref{fig/example1} we can see the resulting text representation for a configuration for words on INEGI bechmark, i.e., the parameters used to transform the original text into the new representation are stemming (\textit{stem}), reduced repeated symbols up to one symbol (\textit{del-d1}), the removal of diacritic (\textit{del-diac}), and coarsening users (\textit{usr}), and negations (\textit{neg}). The final text representation is based on 1-words.

The other example, Figure~\ref{fig/example2}, is a configuration for character 4-gram representation on the same benchmark using the following parameters: the removal of diacritic (\textit{del-diac}), coarsening emoticons (\textit{emo}), coarsening users (\textit{usr}), changing words into lowercase (\textit{lc}), negations (\textit{neg}), and \textsf{TFIDF} is used to weight the tokens, it has no text representation. The final representation is based on character 4-grams, and the underscore symbol is used as space character to separate words and it is part of the token in which it appears.

\begin{figure*}[t!]
\subfigure[An example of configuration for INEGI benchmark for word n-grams] {
    \fbox{
        \begin{minipage}{0.95\textwidth}
        \scriptsize
        {
          \textbf{original text:} \\
         p\'esiiiimo auto :( @autoX fallan frenos y sistema de entretenimiento; no lo compren
         \\
         \textbf{after text transformation:} \\
          pesim aut :( \_user fal fren y sistem de entreten ; lo no\_compr\\
        \textbf{computed 1-word:}\\
       \{pesim, aut, :(, \_user, fal, fren, y, sistem, de, entreten, ;, lo, no\_compr \}
        }
        \end{minipage}
    }
    \label{fig/example1}
}
\subfigure[An example of configuration for INEGI benchmark for q-grams (i.e., $4-$grams) ] {
    \fbox{
        \begin{minipage}{0.95\textwidth}
        \scriptsize
        {
        \textbf{original text:} \\
         ~~p\'esiiiimo auto :( @autoX fallan frenos y sistema de entretenimiento; no lo compren
        } \\
        \textbf{after text transformation:} \\
        ~~pesiiiimo auto \_negativo \_user fallan frenos y sistema de entretenimiento ; lo no\_compren \\
        \textbf{computed 4-grams:}\\
     \{\_pes, pesi, esii, siii, iiii, iiim, iimo, imo\_, mo\_a, o\_au, \_aut, auto, uto\_, to\_\_, o\_\_n, \_\_ne, \_neg, nega, egat, gati, ativ, tivo, ivo\_, vo\_\_, o\_\_u, \_\_us, \_use, user, ser\_, er\_f, r\_fa, \_fal, fall, alla, llan, lan\_, an\_f, n\_fr, \_fre, fren, reno, enos, nos\_, os\_y, s\_y\_, \_y\_s, y\_si, \_sis, sist, iste, stem, tema, ema\_, ma\_d, a\_de, \_de\_, de\_e, e\_en, \_ent, entr, ntre, tret, rete, eten, teni, enim, nimi, imie, mien, ient, ento, nto\_, to\_;, o\_;\_, \_;\_l, ;\_lo, \_lo\_, lo\_n, o\_no, \_no\_, no\_c, o\_co, \_com, comp, ompr, mpre, pren, ren\_ \}

        \end{minipage}
    }\medskip
    \label{fig/example2}
}
\caption{Examples of text representation.}
\label{fig/examples}
\end{figure*}

\section{Benchmarks and Parameter Settings}
\label{sec:problem_parameters}

At this point, we are in the position to analyze the performance of described text representations on sentiment analysis benchmarks. In particular, we test our representations in the task of determining the global polarity ---four polarity levels: positive, neutral, negative, and none (no sentiment)--- of each tweet in two benchmarks.

\begin{table}[t]
  \caption{Datasets details from each competition tested in this work}
  \label{table:benchmarks}
  \begin{center}
  \resizebox{0.8\textwidth}{!}{
  \begin{tabular}{|cl|r rrr|r|}
  \hline
    \multicolumn{2}{|c|} {\bf benchmark} & \multicolumn{4}{c|}{\bf classes}  &     \\ 
   name     & part  &  positive & neutral & negative & none &total\\ \hline
INEGI		& train & 2,908  &  986      & 1,110 &   409 & 5,413  \\
    		& test  & 26,911 &  8,868    & 9,571 & 3,361 & 48,711 \\
            &       &        &           &       &       & 54,124 \\ \hline

TASS'15		& train & 2,884  &  670      & 2,182  &  1,482 & 7,218 \\
    		& test  & 22,233 &  1,305    & 15,844 & 21,416 & 60,798 \\
            &       &        &           &        &        & 68016 \\ \hline
  \end{tabular}
  }
  \end{center}
\end{table}

Table~\ref{table:benchmarks} describes our benchmarks. The INEGI benchmark consists on tweets geo-referenced to Mexico; the data was collected and labeled between 2014 and 2015 by the Mexican Institute of Statistics and Geography (INEGI).
The INEGI's tweets come from the general population without any filtering beyond its geographic location. INEGI benchmark has a total of 54,124 tweets (in the Spanish language). The tagging process of INEGI dataset was conducted through a web application (called pioanalisis\footnote{http://cienciadedatos.inegi.org.mx/pioanalisis/\#/login}, it was designed by the personnel of the Institute). Each tweet was displayed and human tagged it as positive, neutral, negative or none. After this procedure, every tweet was tagged by several humans, the label with major consensus was assigned as a final tagged. We discard tweets being on tie.

On the other hand, our second benchmark is the one used in TASS'15 workshop (Taller de An\'alisis de Sentimientos en la SEPLN)~\cite{tass2015}. Here, the whole corpus contains over $68,000$ tweets, written in Spanish, related to well-known personalities and celebrities of several topics such as politics, economy, communication, mass media, and culture. These tweets were acquired between November 2011 and March 2012. The whole corpus was split into a training set (about 10\%) and test set (remaining 90\%). Each tweet was tagged with its global polarity (positive, negative or neutral) or no polarity at all (four classes in total). The tagging process was done in a semi-automatically way where a baseline machine learning algorithm classifies them, and then all the tagged tweets are manually checked by human experts; for more details of this database construction see ~\cite{tass2015}.
%

We partitioned INEGI in 10\% for training and 90\% for testing, following the setup of TASS'15;
this large test-set pursues the generality of the method. Hereafter, we name the test set as the gold-standard, and we interchange both names as synonyms.
The accuracy is the major score in both benchmarks, again because TASS'15 uses this score as its measure. We also report the macro-F1 score to help to understand the performance on heavily unbalanced datasets, see \ref{table:benchmarks}.

In general, both benchmarks are full of errors, and these errors vary from simple mistakes to deliberate modification of words and syntactic rules.
However, it is worth to mention that INEGI is a collection of an open domain, and moreover, it comes from the general public; then we can see the frequency of misspellings and grammatical errors as a major difference between INEGI and TASS'15.

\begin{figure*}
    \centerline{
    \includegraphics[width=0.5\textwidth]{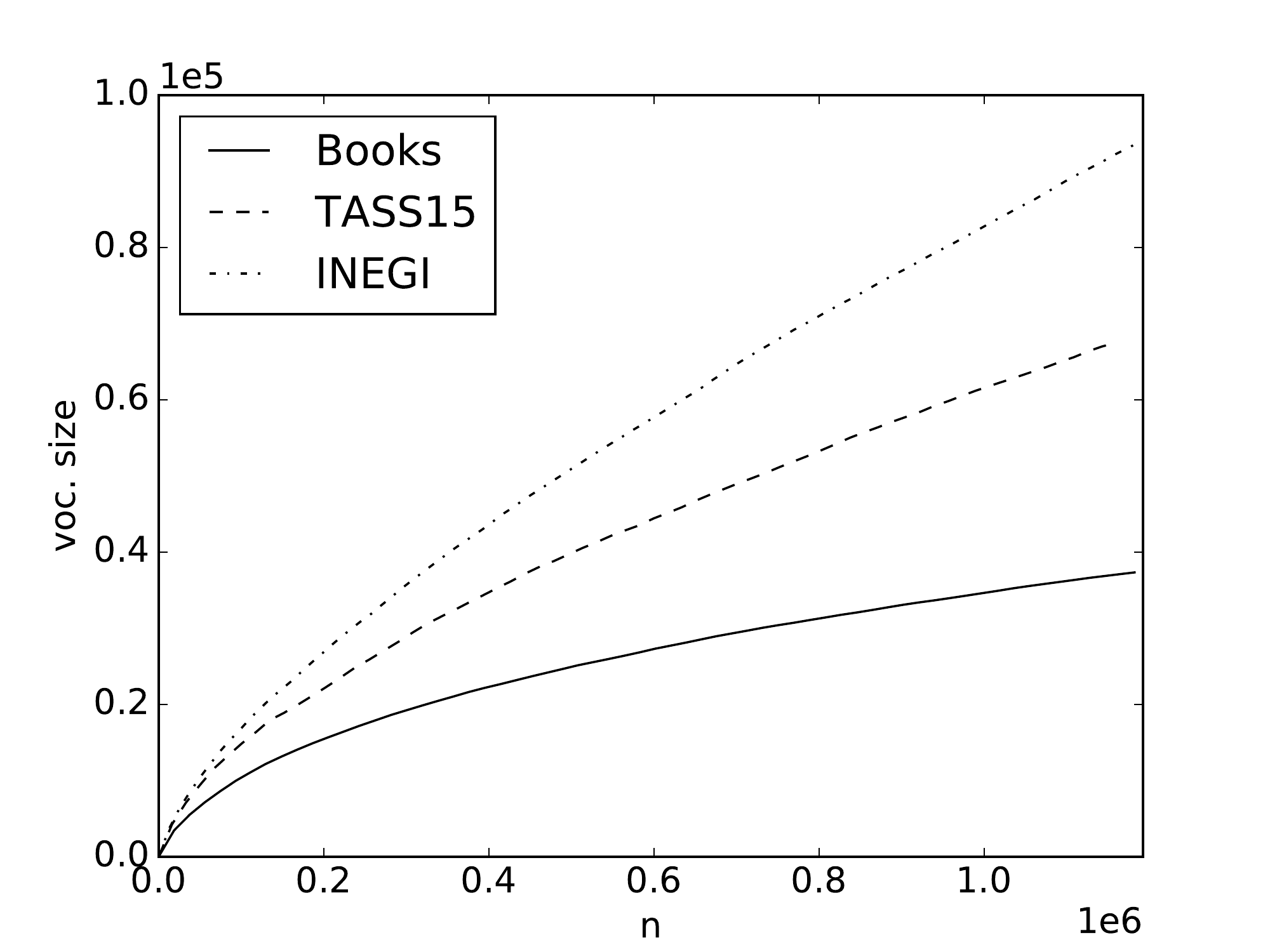}
    \includegraphics[width=0.5\textwidth]{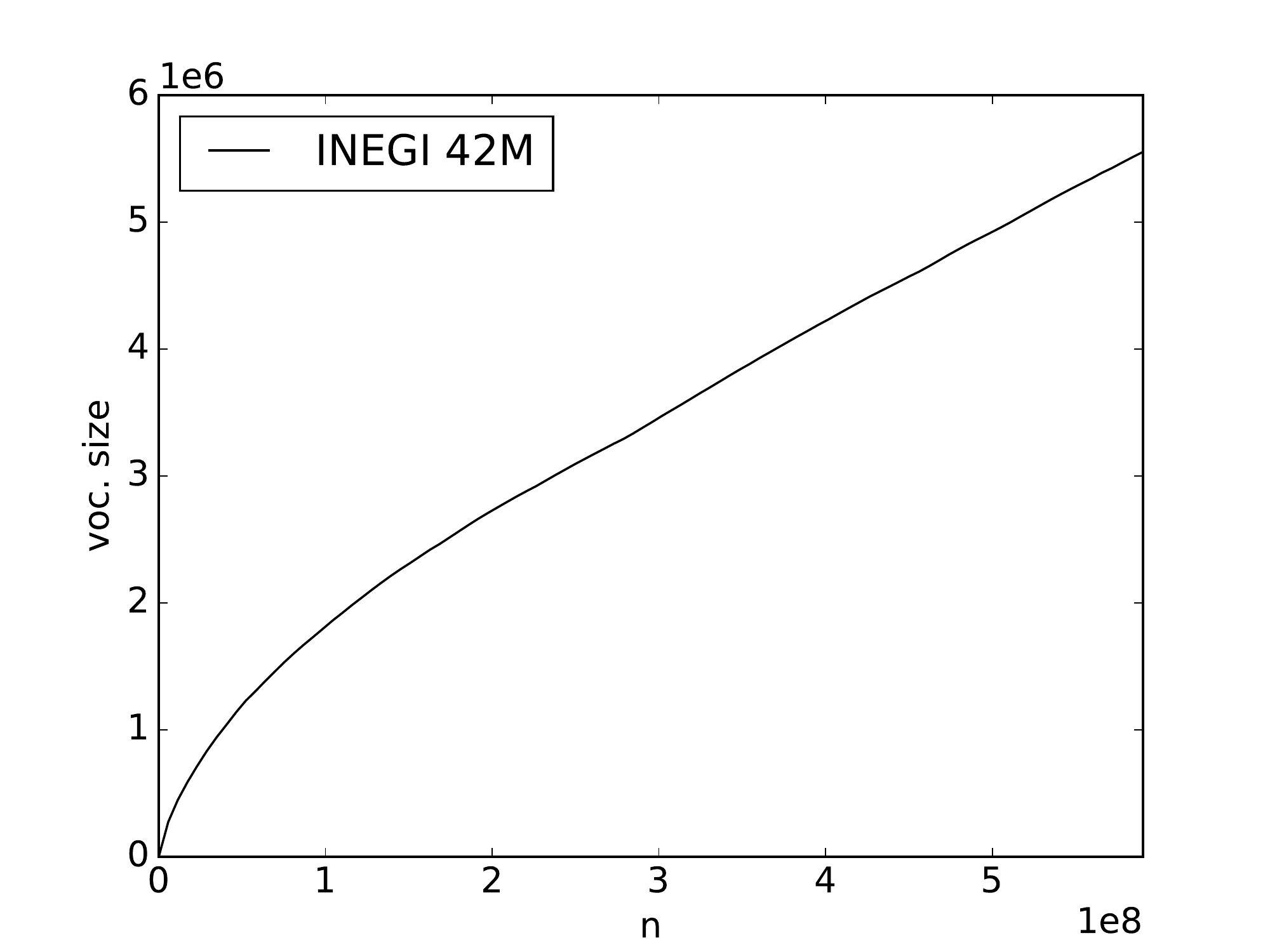}
    }
\caption{On the left, the growth of the vocabulary in our benchmarks and a collection of books from the Gutenberg project. On the right, the vocabulary growth in 42 million tweets.}
\label{fig/heaps}
\end{figure*}

Figure~\ref{fig/heaps} shows the size of the vocabulary as the number of words in the collection increases.
The Heaps' law, \cite{BYRNmir1999}, states that the growth of the vocabulary follows  $O(n^\alpha)$ for $0 < \alpha < 1$, for a document of size $n$. The figure illustrates the growth rate of our both benchmarks, along with a well-written set of documents, i.e., classic Books of the Spanish literature from the Gutenberg project~\cite{gutenberg2016}. The Books collections curve is below than any of our collections;  its growth factor is clearly smaller. The precise values of $\alpha$ for each collection are $\alpha_\textsf{TASS'15} = 0.718$, $\alpha_\textsf{INEGI} = 0.756$, and $\alpha_\textsf{Books} = 0.607$, these values were determined with a regression over the formulae.\footnote{The tweets were slightly normalized removing all URLs and standardizing all characters to lowercase.} There is a significant difference between the three collections, and it corresponds to the high amount of errors in TASS'15, and, the higher one in INEGI.

\subsection{Parameters of the text transformations}
As described in Section \ref{sec:preprocessing} the different text transformation methods explored in this research. Table~\ref{tab/parameters} complements this description by listing the different values these transformations have. From the table, it can be observed that most parameters are either the use or absence of the particular transformation with the exceptions n-words and $q$-grams.

Based on the different values of the parameters, we can count the number of different text transformation which is $7 \times 2^{15} = 229,369$ configurations (the constant $7$ corresponds to the number of tokenizers). Evaluating all these setups, for each benchmark, is computationally expensive. Also, we perform the same exhaustive in the test set to compare the achieved result and the best possible under our approach. Along with these experiments, we also evaluate a number of experiments to prove and compare a series of improvements. In the end, we evaluated close to one million configurations.
For instance, using an Intel(R) Xeon(R) CPU E5-2630 v2 @ 2.60GHz workstation, we need $\sim\!\!12$ minutes in average for a single configuration, running on a single core. Therefore, it needs roughly 24 years of computing time. Nonetheless, we used a small cluster to compute all configurations in some weeks. Notice that the time of determining the part-of-the-speech, needed by parameters {\em stem} and {\em lem}, is not reported since it was executed only once for all texts and loaded from a cache whenever is needed. The lemmatization step needs close to $56$ minutes to transform the INEGI dataset in the same hardware.

\begin{table*}
\caption{Parameter list and a brief description of their functionality}
\label{tab/parameters}
\centering
\resizebox{!}{0.48\textheight}{
\begin{tabular}{c@{~~}c@{~~}p{0.7\textwidth}} \hline

\multicolumn{3}{c}{\bf weighting schemes / removing common words} \\ \hline
\bf name & \bf values & \bf \hspace{3cm} description \\

tfidf  & yes, no & After the text is represented as a bag of words, it determines if the vectors are weighted using the \textsf{TFIDF} scheme. If it is {\em no} then the term frequency in the text is used as weight.\\
del-sw & yes, no & Determines if the stopwords are removed. It is related to \textsf{TFIDF} in the sense that a proper weighting scheme assigns a low weight for common words. \\

\hline
\multicolumn{3}{c}{\bf morphological reductions} \\ \hline
\bf name & \bf values & \bf \hspace{3cm} description \\
lem    & yes, no & Determines if words sharing a common root are replaced by its root. \\
stem   & yes, no & Determines if words are stemmed. \\

\hline
\multicolumn{3}{c}{\bf transformations based on removing or replacing substrings} \\ \hline
\bf name & \bf values & \bf \hspace{3cm} description \\
del-punc & yes, no & The punctuation symbols are removed if {\em del-punc} is {\em yes}, they are left untouched otherwise. \\
del-ent  & yes, no & Determines if entities are removed in order to generalize the content of the text. \\
del-d1   & yes, no & If it is enabled then the sequences of repeated symbols are replaced by a single occurrence of the symbol. \\
del-d2   & yes, no & If it is enabled then the repeated sequences of two symbols are replaced by a single occurrence of the sequence. \\
del-diac & yes, no & Determines if diacritic symbols, e.g., accent symbols, should be removed from the text. \\

\hline
\multicolumn{3}{c}{\bf coarsening transformations} \\ \hline
\bf name & \bf values & \bf \hspace{3cm} description \\
emo      & yes, no & Emoticons are replaced by its expressed emotion if it is enabled. \\
num      & yes, no & Determines if numeric words are replaced by a common identifier. \\
url      & yes, no & Determines if URLs are left untouched or replaced by a unique url identifier. \\
usr      & yes, no & Determines if users mentions are replaced by a unique user identifier. \\
lc       & yes, no & Letters are normalized to be lowercase if it is enabled \\

\hline
\multicolumn{3}{c}{\bf handling negation words} \\ \hline
\bf name & \bf values & \bf \hspace{3cm} description \\
neg    & yes, no & Determines if negation operators in the text are normalized and directly connected with the modified object. \\

\hline
\multicolumn{3}{c}{\bf tokenizing the transformation} \\ \hline
\bf name & \bf values        & \bf \hspace{3cm} description \\
n-words    & $\{1, 2\}$      & Determines the number of words used to describe a token. \\
q-grams  & $\{3,4,5,6,7\}$ & Determines the length in characters of the $q$-grams ($q$). \\
\hline
\end{tabular}
}
\end{table*}


\section{Experimental Analysis}
\label{sec:results}
This section is devoted to describe and analyze the performance of the configuration space, provide the sufficient experimental evidence to prove that $q$-gram tokenizers are better than $n$-words, at least under the sentiment analysis domain in Spanish. Furthermore, we also provide the experimental analysis for the combination of tokenizers, which improves the whole performance without moving too far from our text classifier structure.

We use both training and test datasets in our experiments. The performance on the training set is computed using 5-fold cross validation, and the performance on test set is computed directly on the gold-standard. As previously described, training and test are disjoint sets, see Table~\ref{table:benchmarks} for details of our benchmarks.
As mentioned, the classifier was fixed to be SVM; we use the implementation from the Scikit-learn project~\cite{scikit-learn} using a linear kernel. We use the default parameters of the library; no additional tuning was performed in this sense.

\subsection{A Performance Comparison of $n$-words and $q$-grams}

\begin{figure*}[t!]
\subfigure[Performance for n-words in training subset] {
    \includegraphics[width=0.5\textwidth]{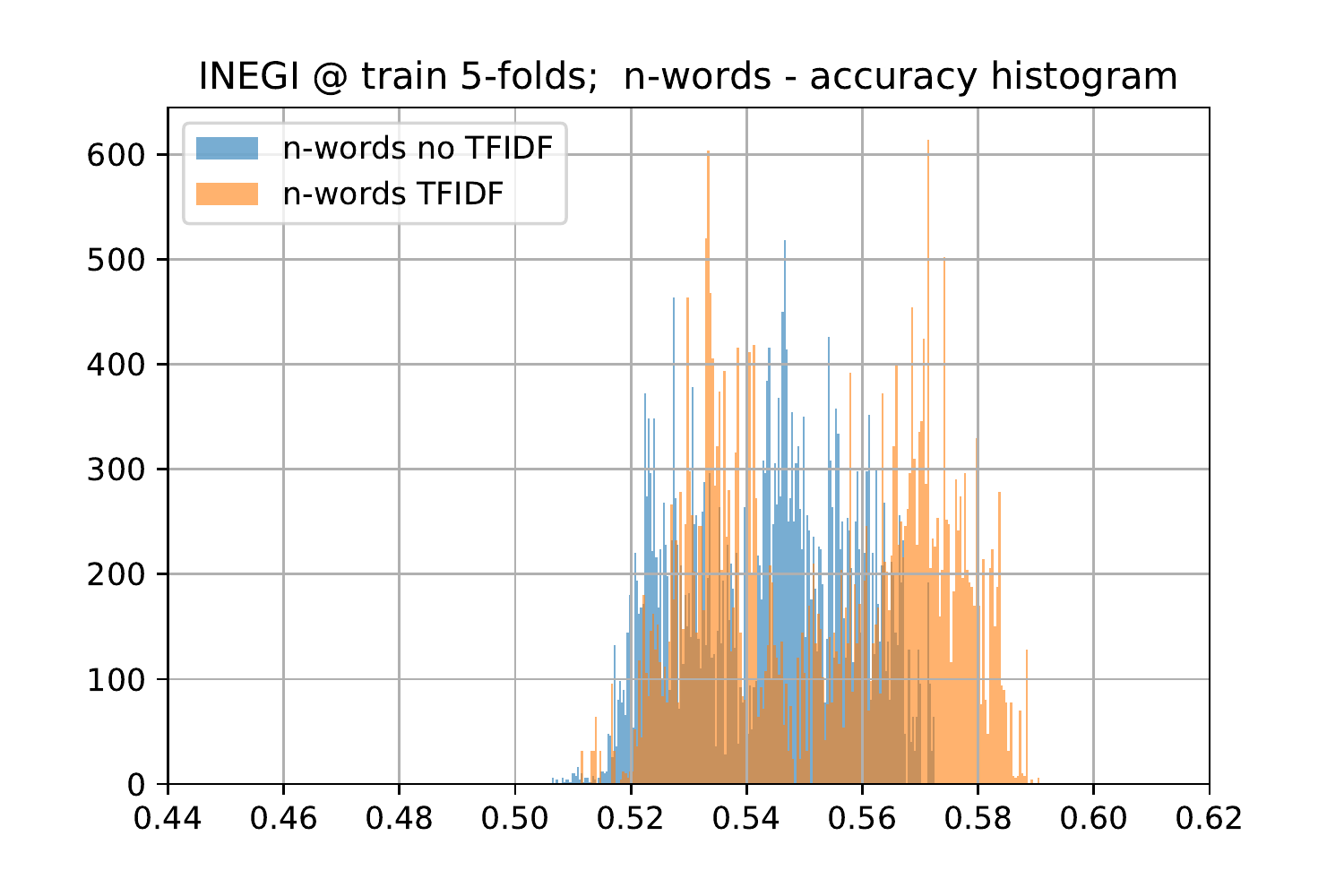}
    \label{fig/inegi-nwords/train}
}
\subfigure[Performance for $q$-grams in training subset]{
    \includegraphics[width=0.5\textwidth]{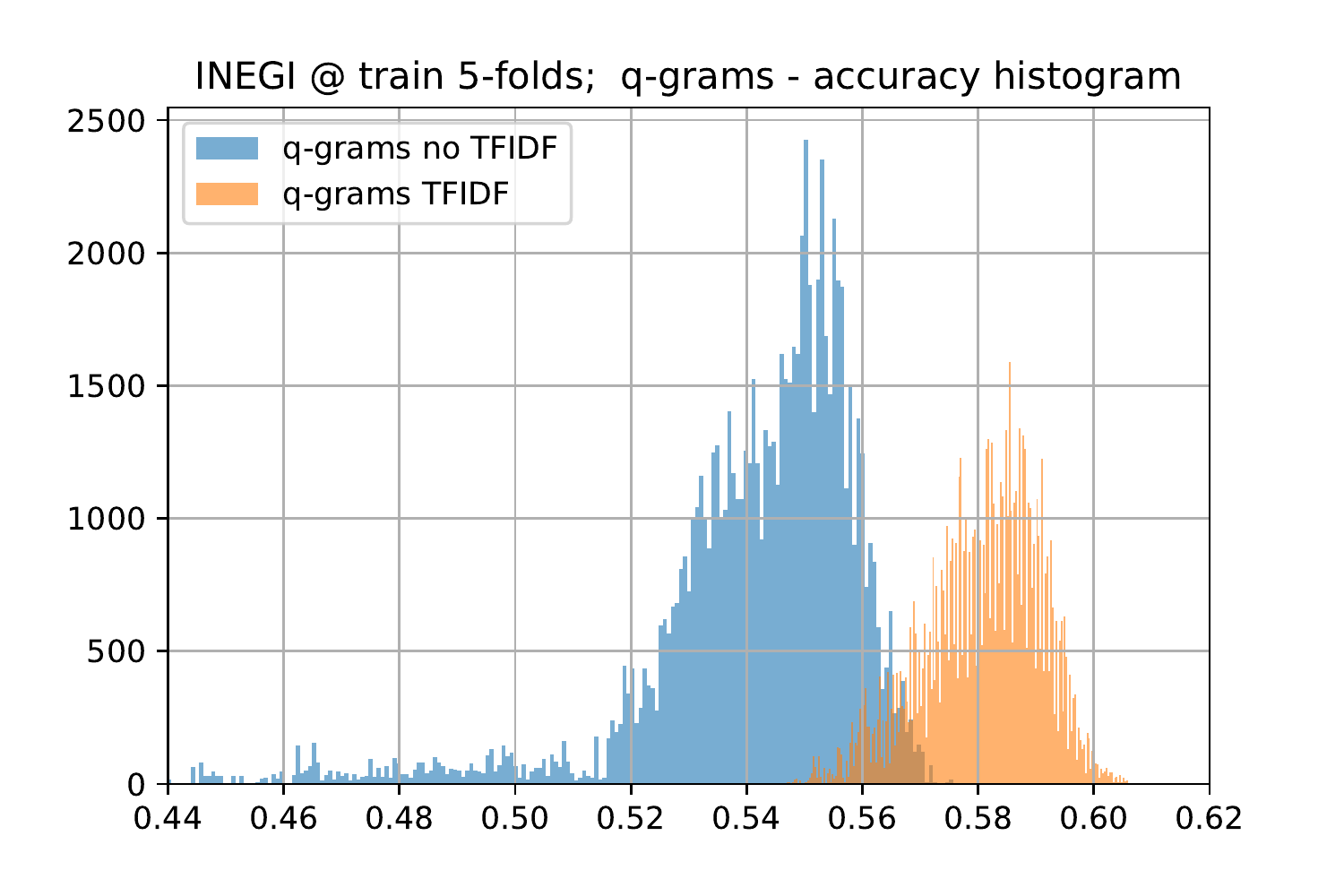}
    \label{fig/inegi-qgrams/train}
}

\subfigure[Performance for n-words in gold standard]{
    \includegraphics[width=0.5\textwidth]{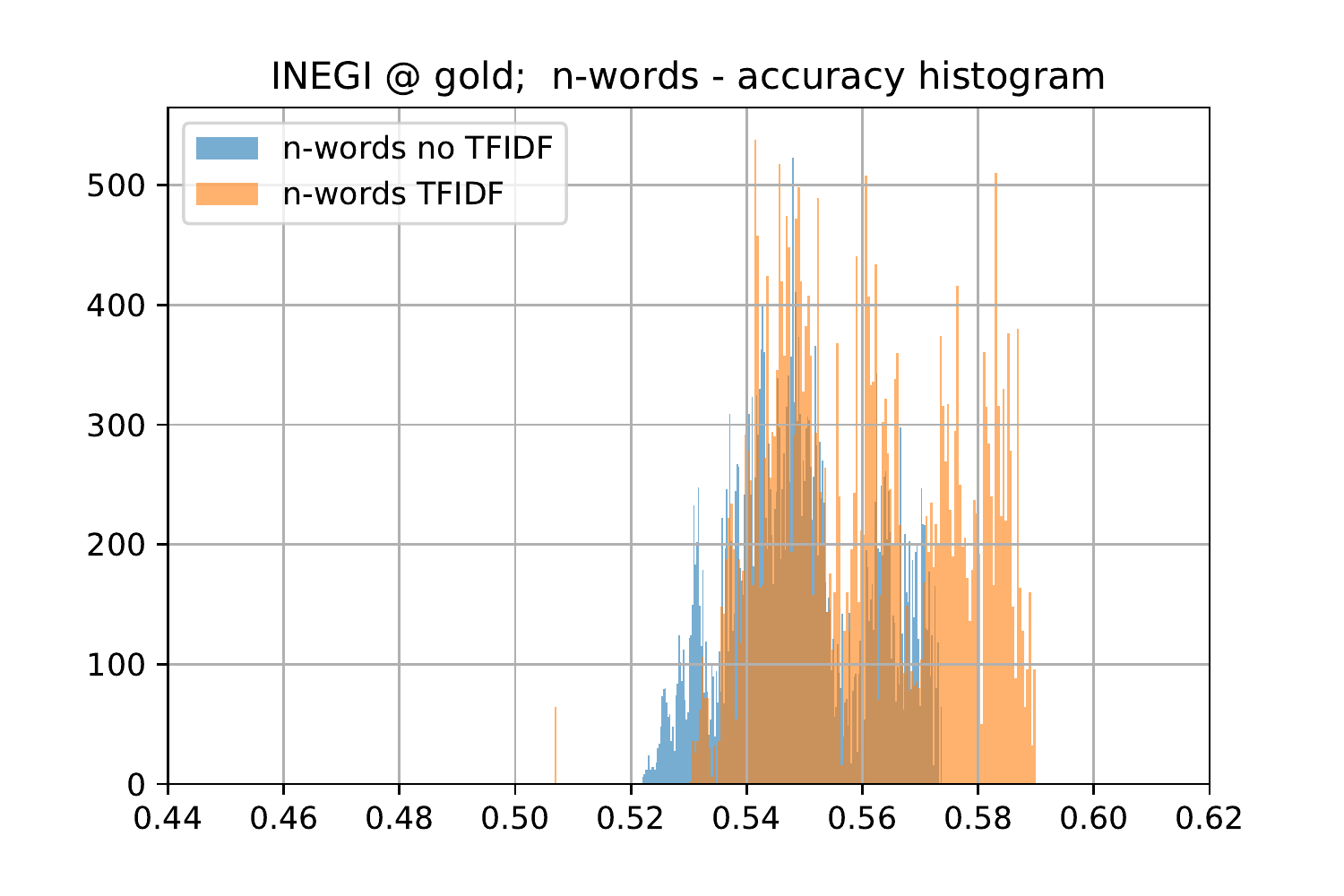}
    \label{fig/inegi-nwords/gold}
}
\subfigure[Performance for $q$-grams in gold standard]{
    \includegraphics[width=0.5\textwidth]{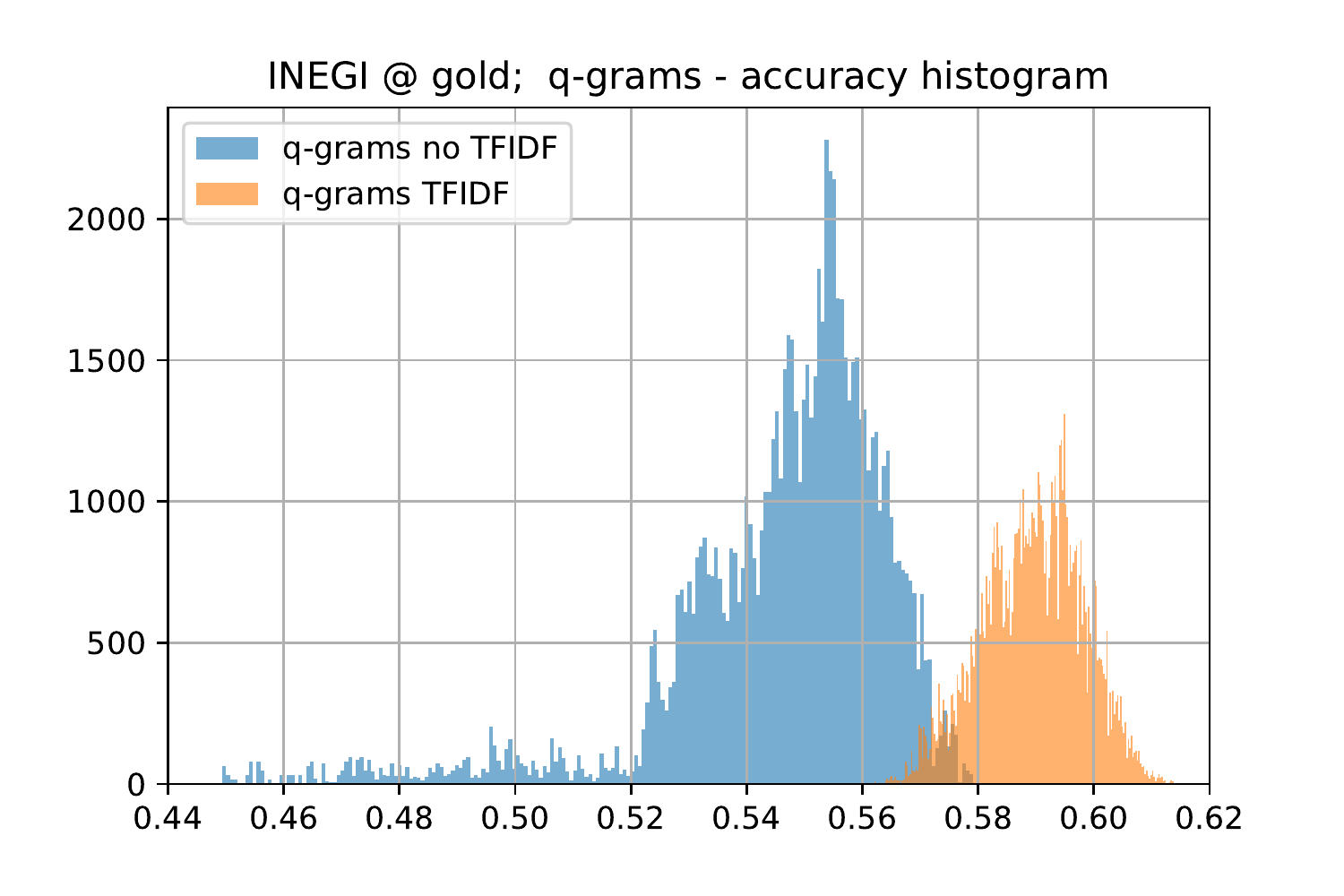}
    \label{fig/inegi-qgrams/gold}
}
\caption{Accuracy's histogram, by tokenizer's class, for the INEGI benchmark. The performance on the training set was computed with 5-folds. We select to divide each figure to show  the effect of \textsf{TFIDF}, which it is essencial for $q$-grams's performance.}
\label{fig/inegi-performance}
\end{figure*}

Figure~\ref{fig/inegi-performance} shows the histogram of accuracies for our configuration-space in both training and test partitions. Figures~\ref{fig/inegi-nwords/train} and \ref{fig/inegi-nwords/gold} show the performance of configurations with $n$-words as tokenizer (unigrams and bigrams), for training and test datasets respectively. It is possible to see that the form is preserved, and also that \textsf{TFIDF} configurations can perform slightly better than those using only the term frequency. However, the accuracy range being shared by both kinds of configurations is large.

In contrast, Figure~\ref{fig/inegi-qgrams/train} shows the performance of configurations with $q$-grams as tokenizers. Here, the improvement of the \textsf{TFIDF} class is more significant than those configurations not using \textsf{TFIDF}; also, the performance achieved by  the $q$-grams with \textsf{TFIDF} is consistently better than the performance of the all $n$-word configurations in our space. This is also valid for the test dataset, see Figure~\ref{fig/inegi-qgrams/gold}.

\begin{figure*}[t!]
\subfigure[Performance without \textsf{TFIDF} on  training subset] {
    \includegraphics[width=0.5\textwidth]{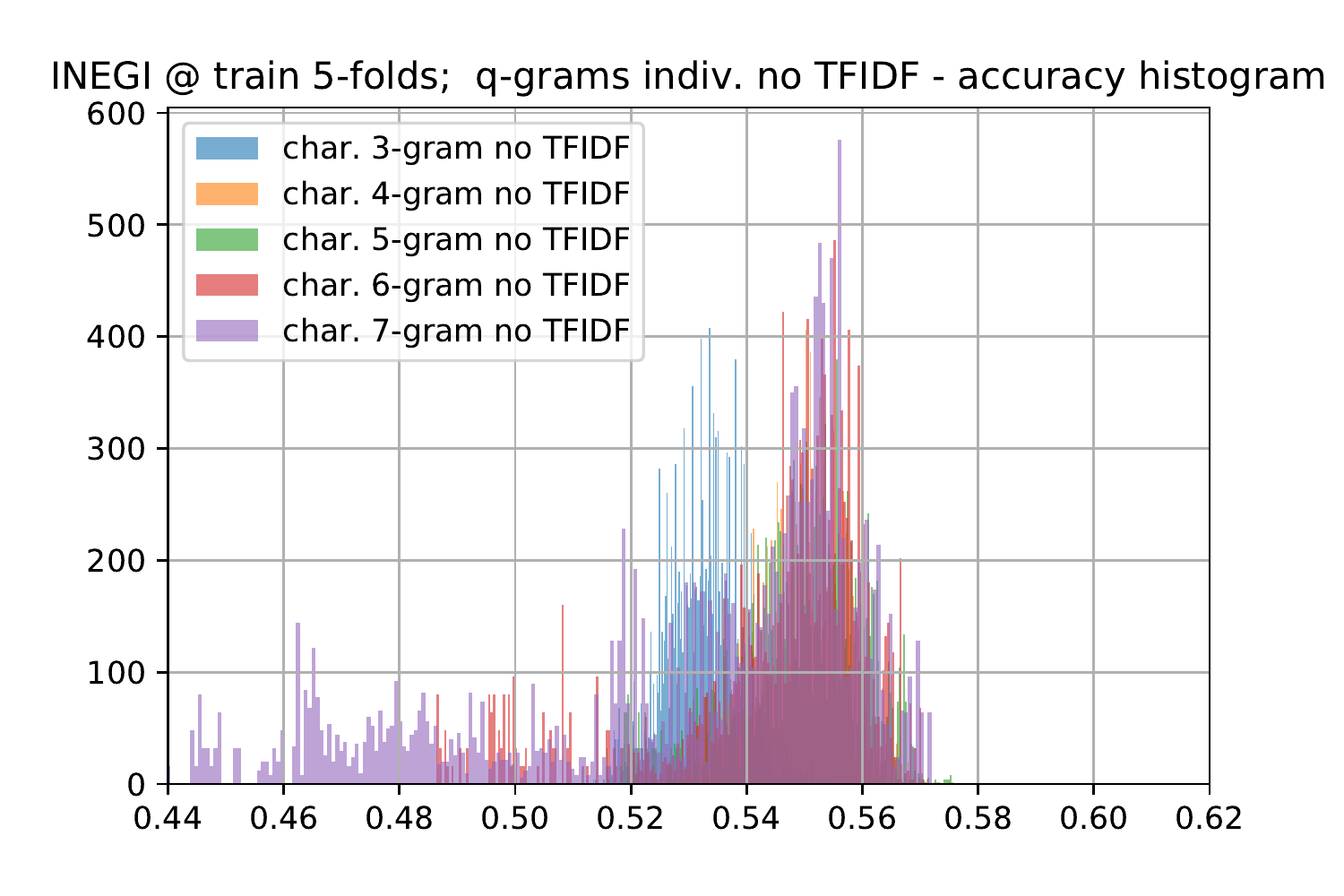}
    \label{fig/inegi/qgrams/train/no-tfidf}
}
\subfigure[Performance with \textsf{TFIDF} on training subset]{
    \includegraphics[width=0.5\textwidth]{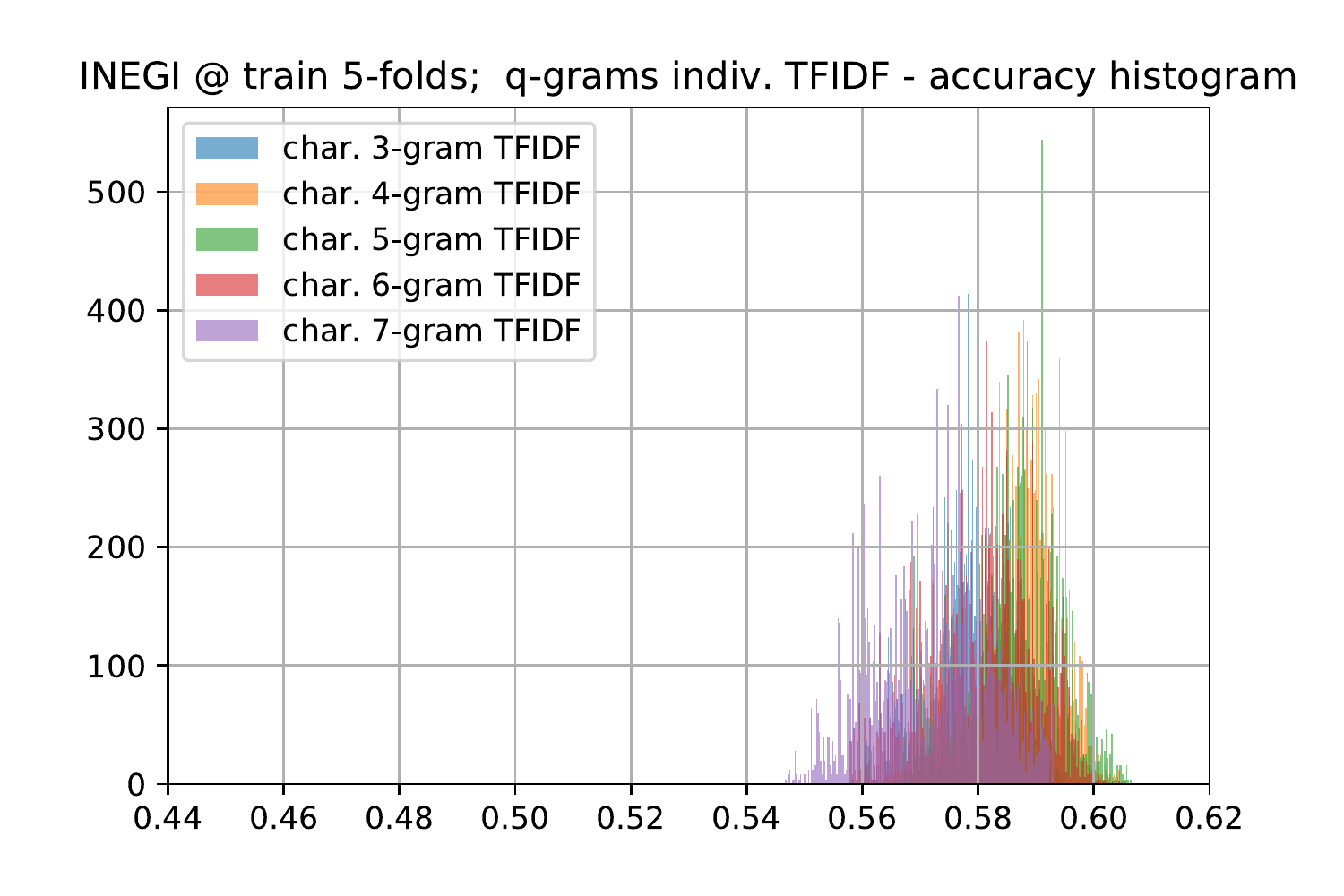}
    \label{fig/inegi/qgrams/train/tfidf}
}

\subfigure[Performance without \textsf{TFIDF} on the gold-standard] {
    \includegraphics[width=0.5\textwidth]{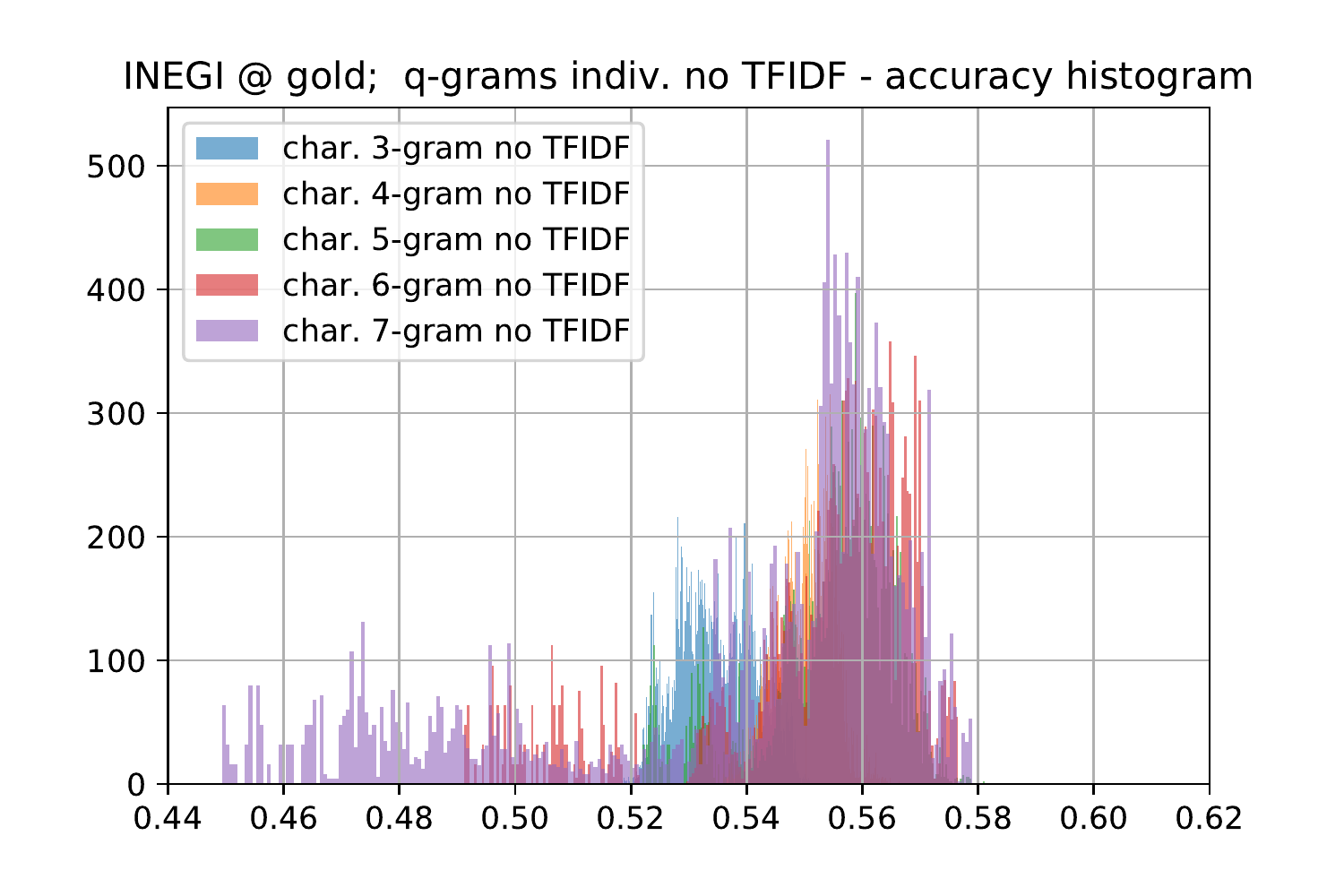}
    \label{fig/inegi/qgrams/gold/no-tfidf}
}
\subfigure[Performance with \textsf{TFIDF} in the gold standard]{
    \includegraphics[width=0.5\textwidth]{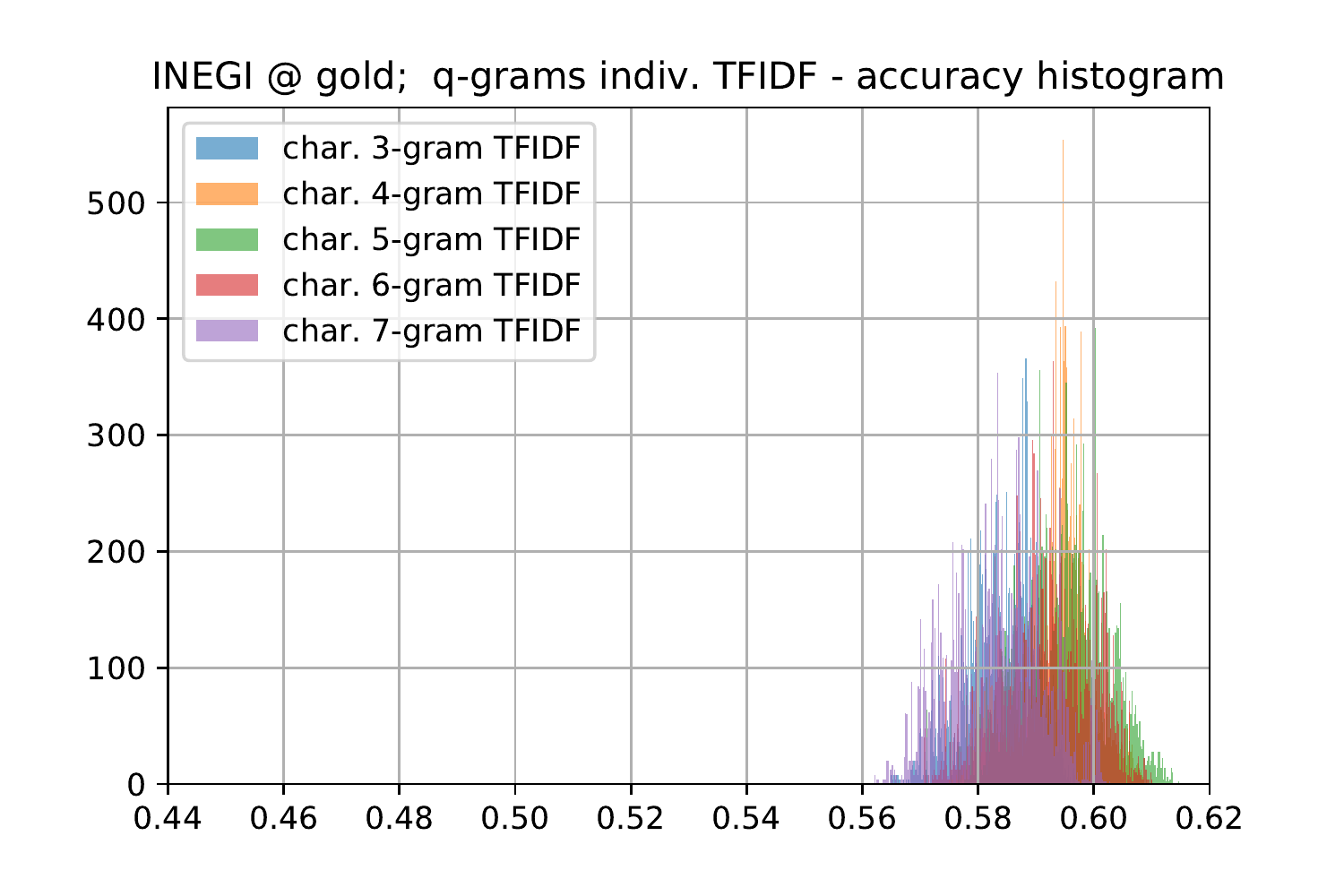}
    \label{fig/inegi/qgrams/gold/tfidf}
}
\caption{Accuracy's histogram for q-gram configurations in the INEGI benchmark. As before, the performance on the training set was computed with 5-folds.}
\label{fig/inegi/qgrams}
\end{figure*}

Figure~\ref{fig/inegi/qgrams} shows the performance of INEGI on configurations using $q$-grams as tokenizers. On the left, Figures~\ref{fig/inegi/qgrams/train/no-tfidf} and \ref{fig/inegi/qgrams/gold/no-tfidf} show the performance of configurations without \textsf{TFIDF}.
In train, the best performance is close to 0.57, and less than 0.58 in the test set. The best performing tokenizer is $7$-grams. When \textsf{TFIDF} is allowed, Tables~\ref{fig/inegi/qgrams/train/tfidf} and \ref{fig/inegi/qgrams/gold/tfidf}, the best performances are achieved, in both training and test, close to 0.61 in the training set and higher in the gold-standard.
The best configurations are those with $5$-grams and $6$-grams. The $5$-grams is consistently better, it achieves accuracy values of 0.6065 and 0.6148 for training and test sets, respectively.


\subsubsection{Performance on the TASS'15 benchmark}
\begin{figure*}[t!]

\subfigure[Performance for n-words in training subset] {
    \includegraphics[width=0.5\textwidth]{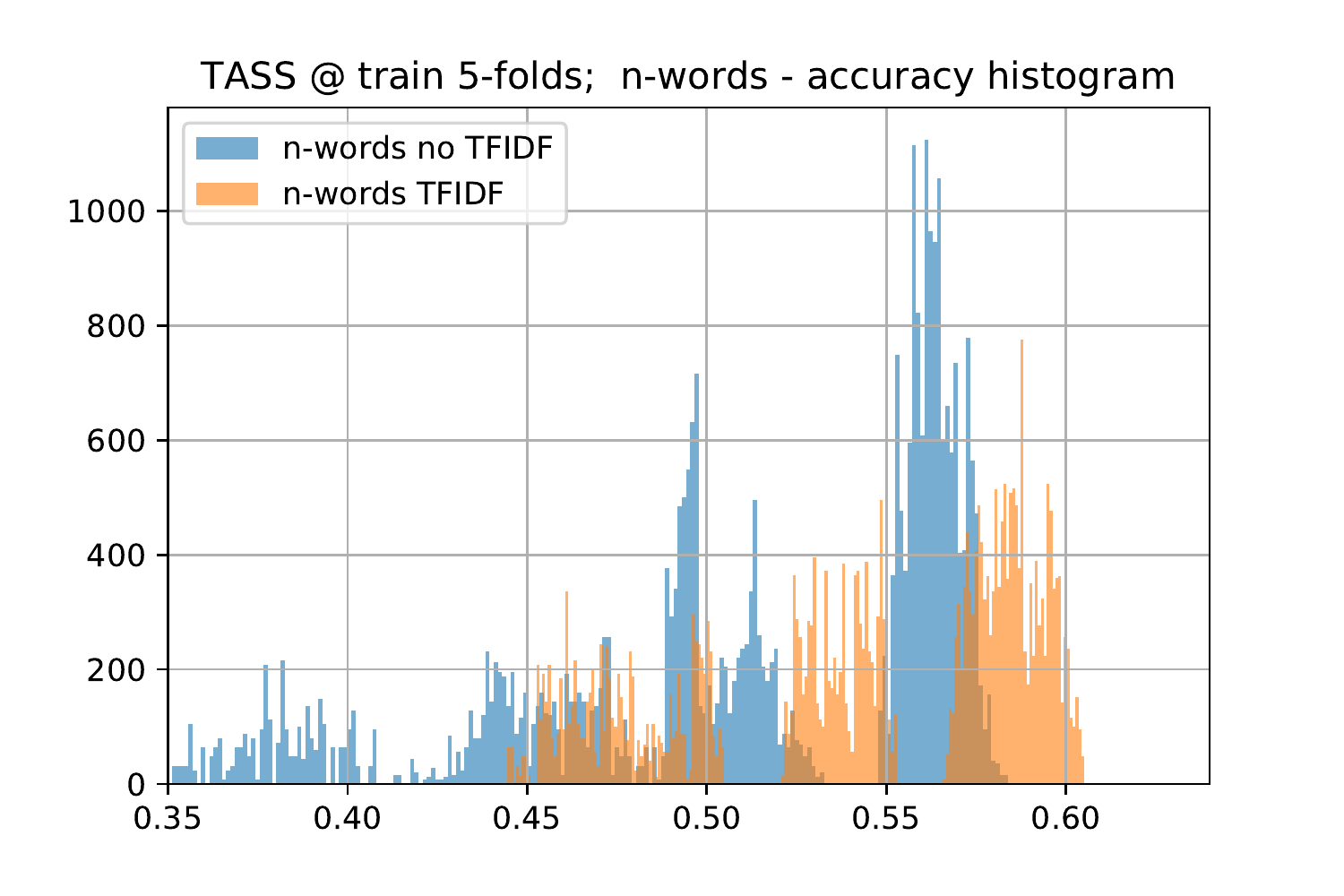}
    \label{fig/tass-nwords/train}
}
\subfigure[Performance for $q$-grams in training subset]{
    \includegraphics[width=0.5\textwidth]{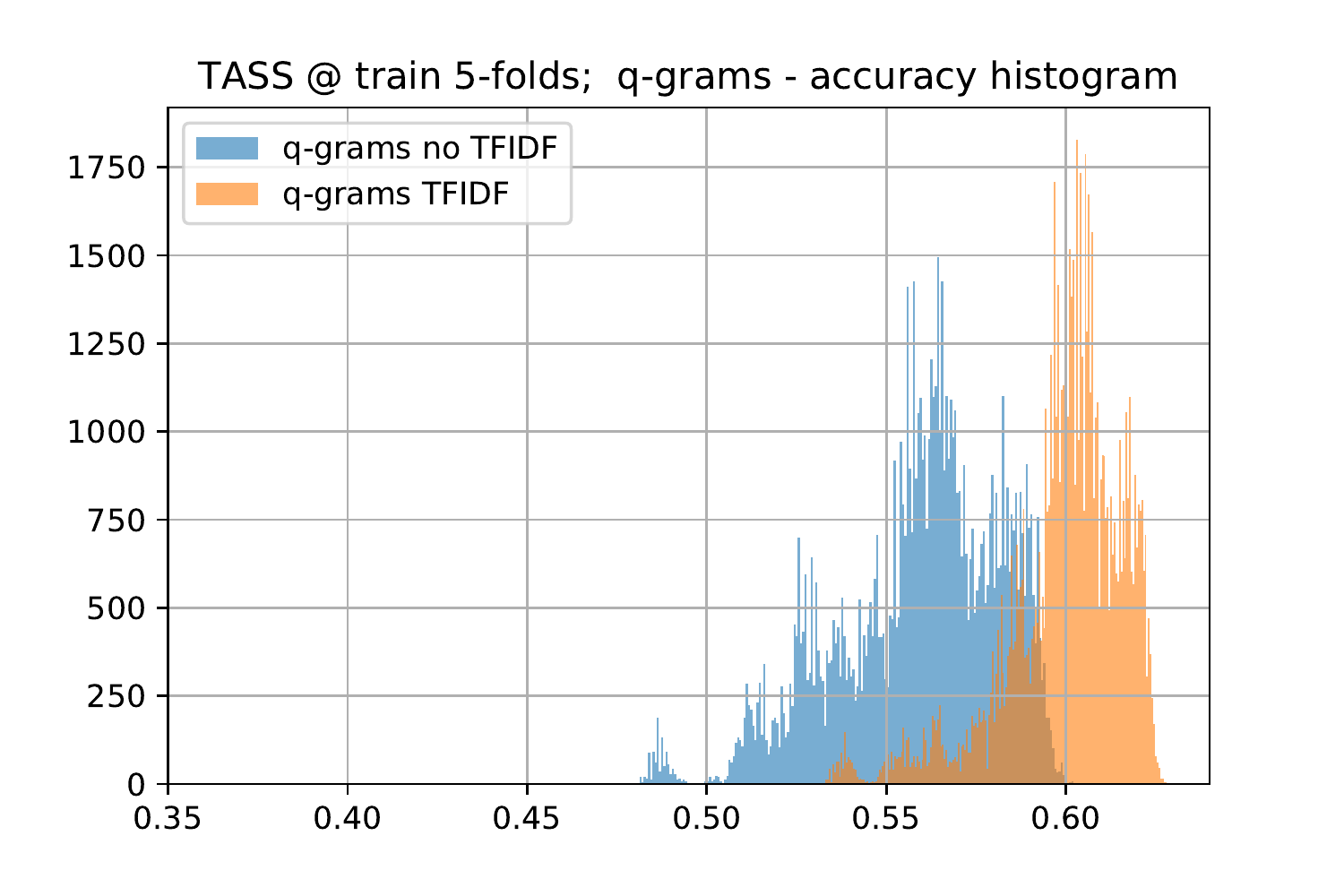}
    \label{fig/tass-qgrams/train}
}

\subfigure[Performance for n-words in gold standard]{
    \includegraphics[width=0.5\textwidth]{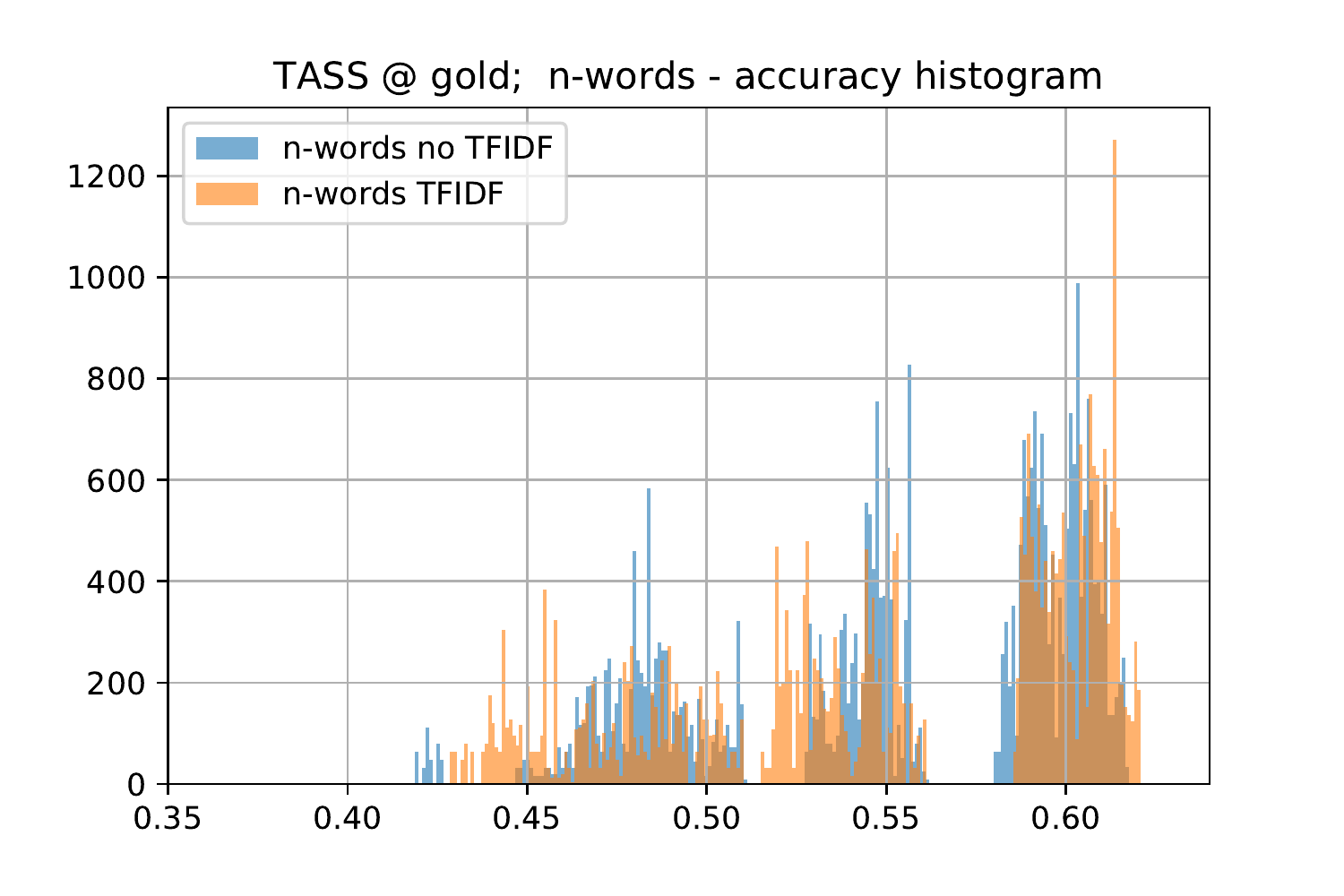}
    \label{fig/tass-nwords/gold}
}
\subfigure[Performance for $q$-grams in gold standard]{
    \includegraphics[width=0.5\textwidth]{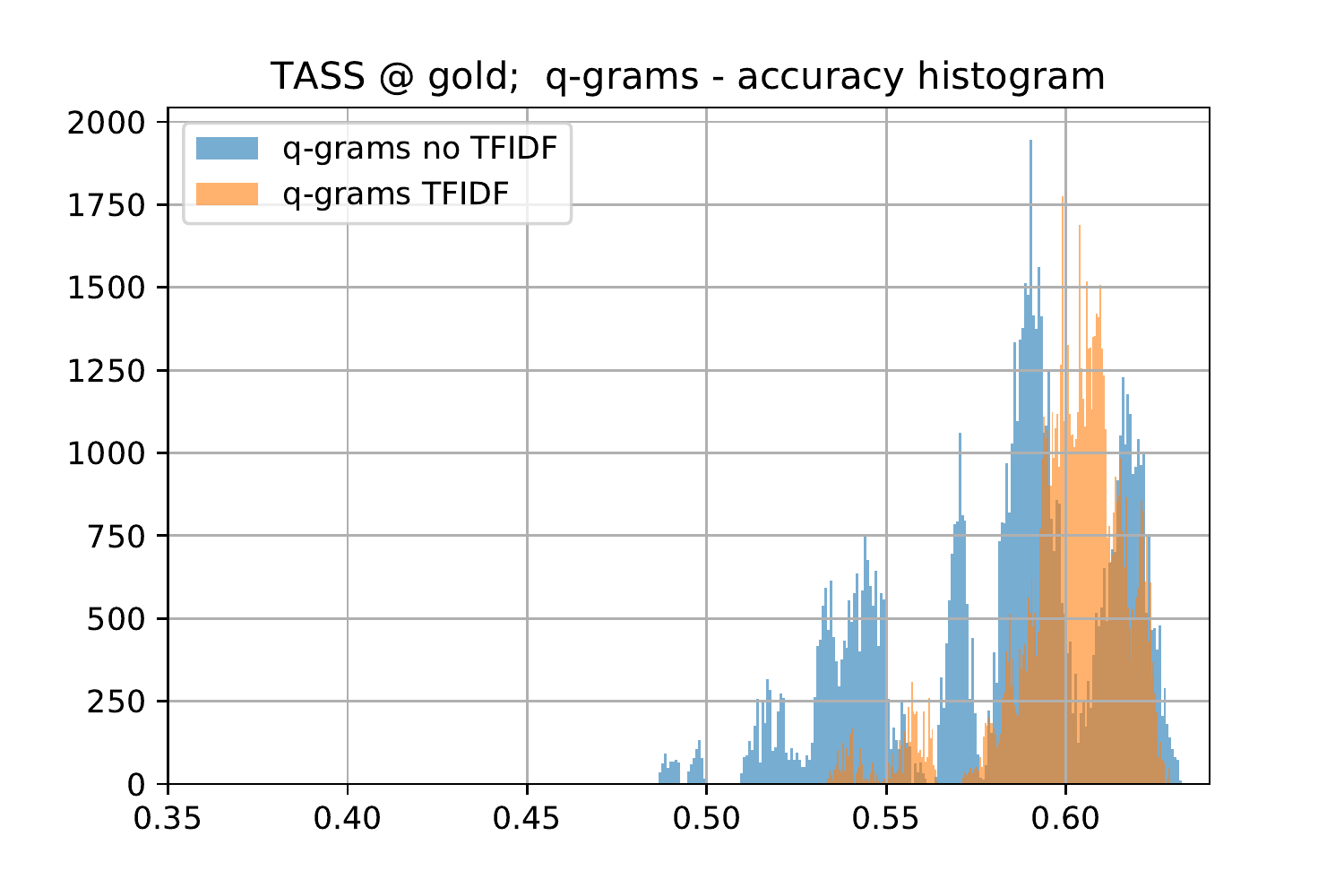}
    \label{fig/tass-qgrams/gold}
}
\caption{Accuracy's histogram, by tokenizer's class, for the TASS benchmark. The performance on the training set was computed with 5-folds. We select to divide each figure to show  the effect of \textsf{TFIDF}, which it is essencial for $q$-grams's performance.}
\label{fig/tass-performance}
\end{figure*}

The performance on TASS'15 is similar to that found in the INEGI benchmark; however, TASS'15 shows a higher sparsity of the accuracy along the range on $n$-words, ranging from 0.35 to close than 0.61.
In the training set, the best performances are achieved using \textsf{TFIDF}.

The best configurations are those using $q$-grams, as depicted in Figure~\ref{fig/tass-qgrams/train} and \ref{fig/tass-qgrams/gold}, where accuracy values achieve close to 0.63 in both training and test sets. In contrast to INEGI and the training set of TASS'15, the best performing $q$-gram tokenizer has no \textsf{TFIDF}, however the configurations with \textsf{TFIDF} are tightly concentrated which means that is more easy to pick a good configuration under a random selection, or by the insight of an expert.

\begin{figure*}[th!]
\subfigure[Performance without \textsf{TFIDF} in training subset] {
    \includegraphics[width=0.5\textwidth]{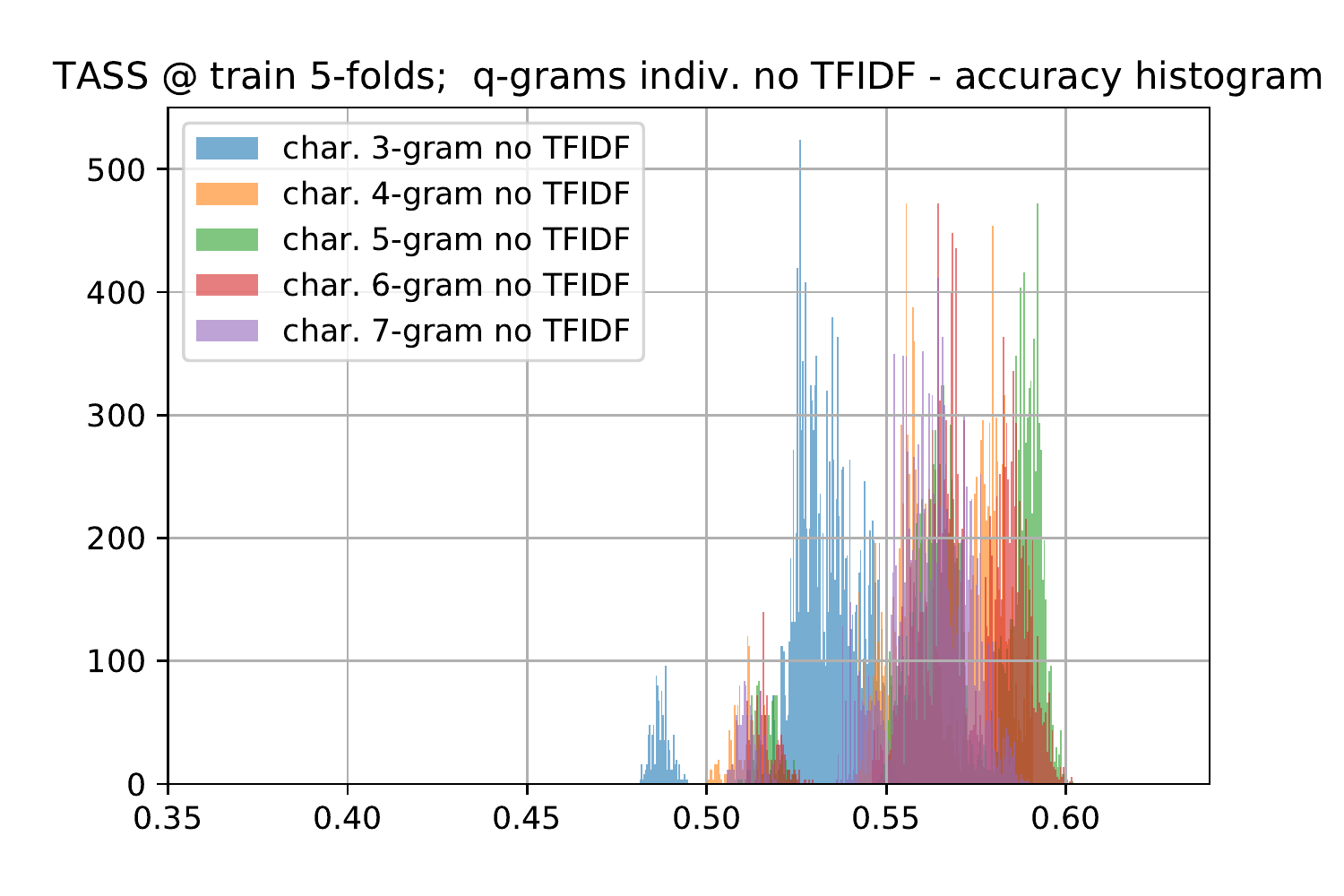}
    \label{fig/tass-all-qgrams/train/no-tfidf}
}
\subfigure[Performance with \textsf{TFIDF} in training subset]{
    \includegraphics[width=0.5\textwidth]{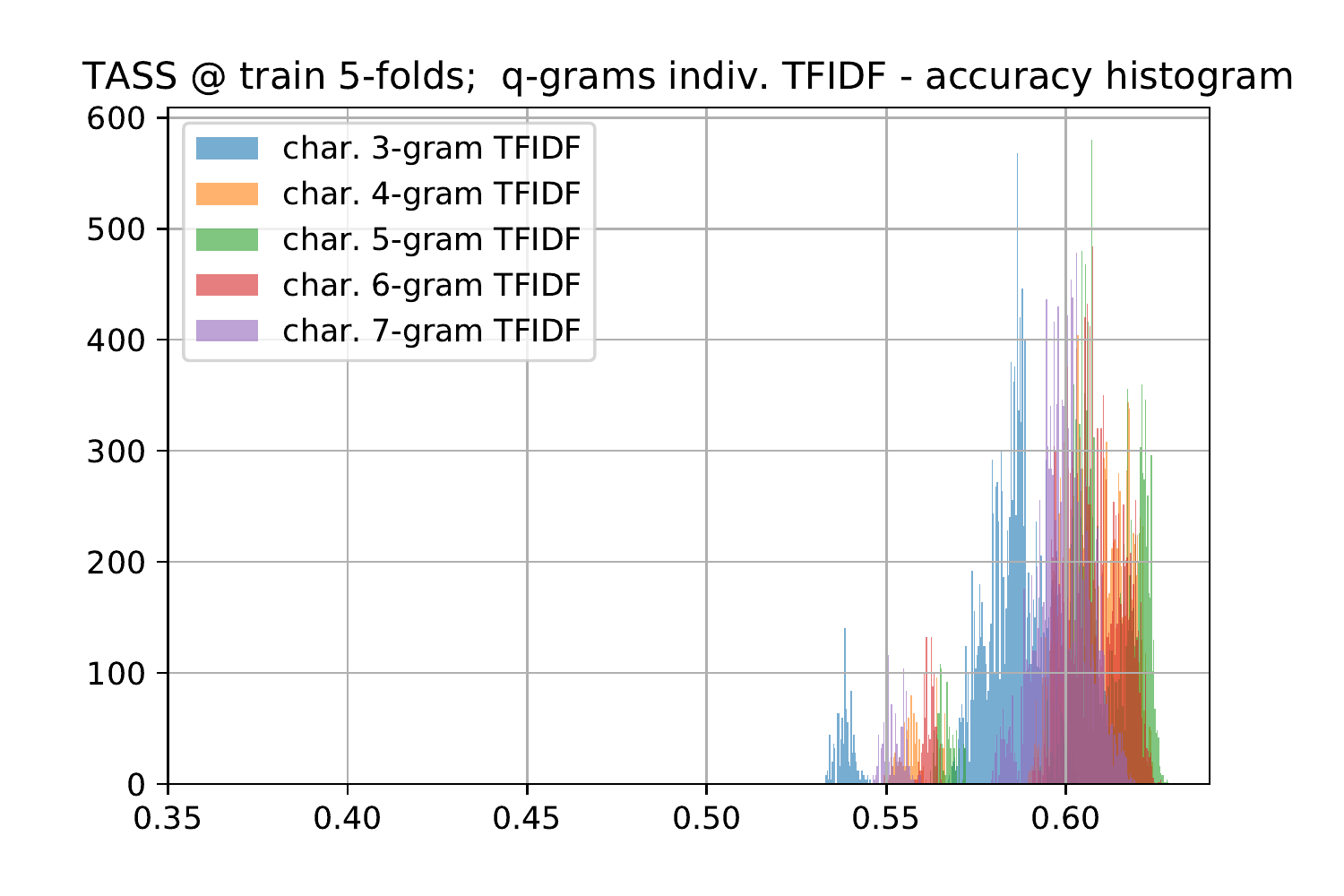}
    \label{fig/tass-all-qgrams/train/tfidf}
}

\subfigure[Performance without \textsf{TFIDF} in the gold standard] {
    \includegraphics[width=0.5\textwidth]{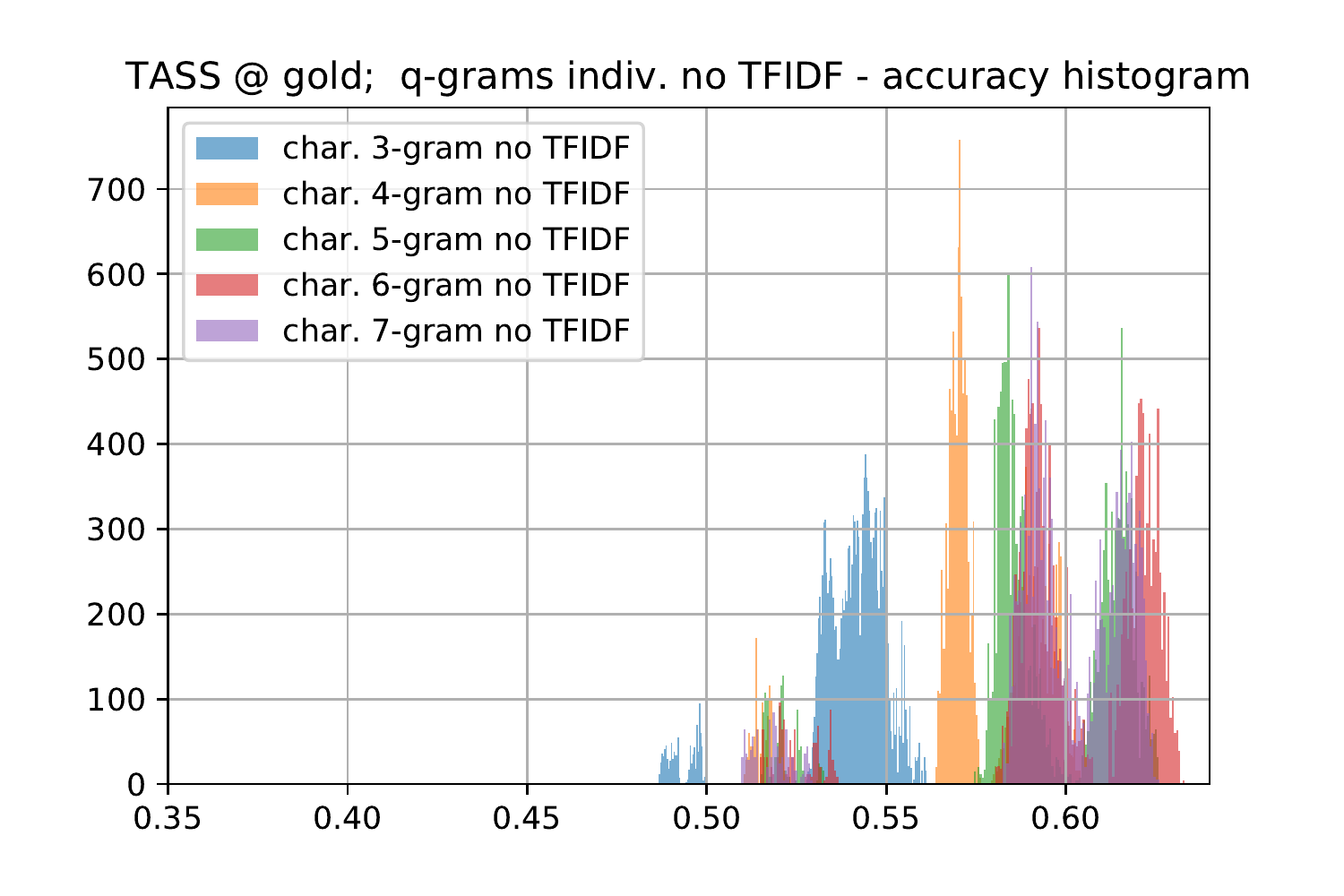}
    \label{fig/tass-all-qgrams/gold/no-tfidf}
}
\subfigure[Performance with \textsf{TFIDF} in the gold standard]{
    \includegraphics[width=0.5\textwidth]{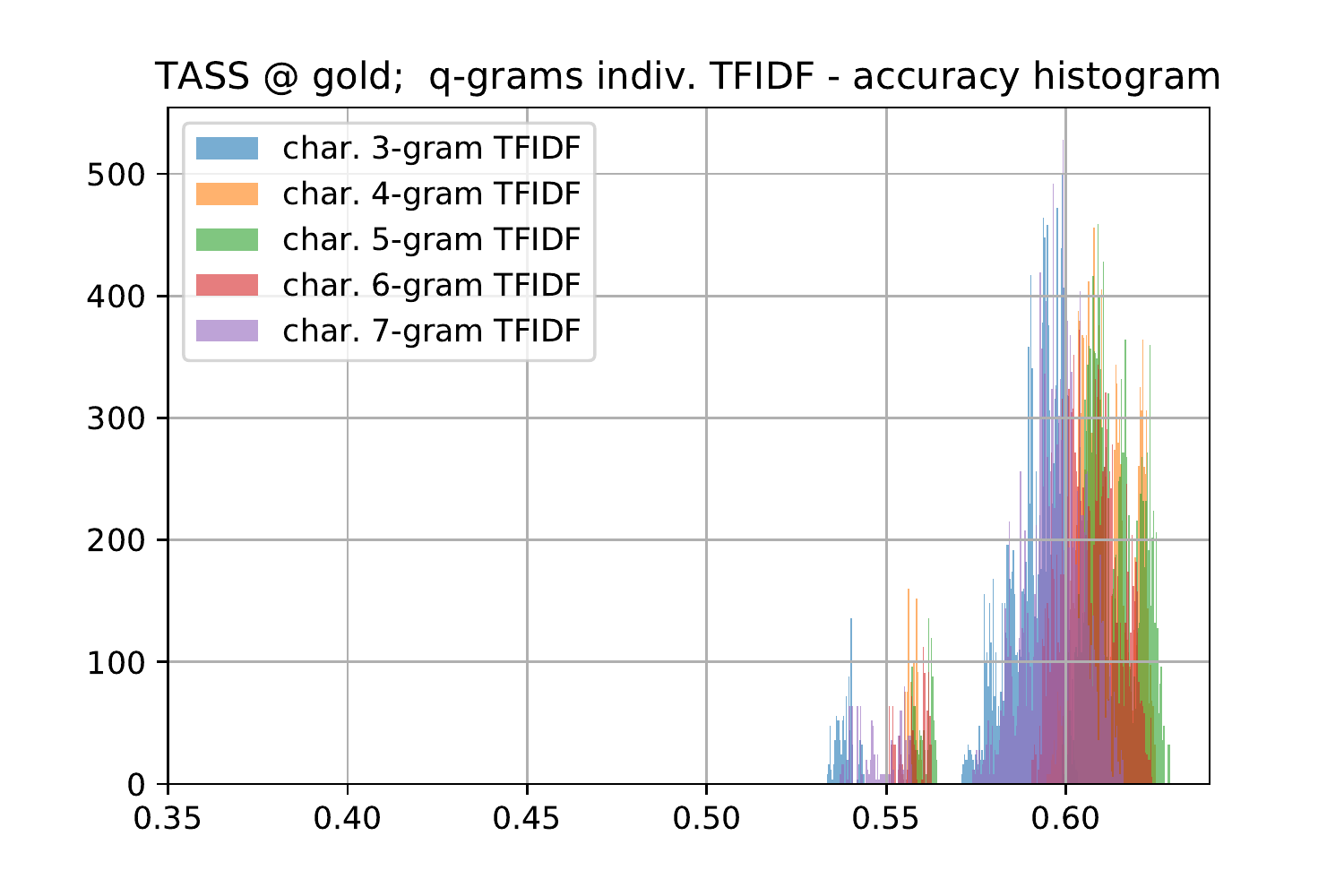}
    \label{fig/tass-all-qgrams/gold/tfidf}
}
\caption{Accuracy's histogram for q-gram configurations in the TASS benchmark. As before, the performance on the training set was computed with 5-folds.}
\label{fig/tass/qgrams}
\end{figure*}

Figure~\ref{fig/tass/qgrams} shows a finer analysis of the performance of $q$-grams tokenizers in TASS'15. We can observe that $5$-grams appear as the best in the training set and in the gold-standard with \textsf{TFIDF}, but the best performing configuration uses $6$-grams tokenizers and no \textsf{TFIDF};
please note that \textsf{TFIDF} has the best accuracy on the training set, so we have not way to know this behaviour without testing all possible configurations in the gold-standard.
Also, the difference between the best \textsf{TFIDF} and the best no-\textsf{TFIDF} configurations is of around 0.005; that is quite small to discard the current bias that suggest to use \textsf{TFIDF} configurations.

\subsection{Top-$k$ Analysis}
This section focus on the structural analysis of the best $k$ configurations (based on the accuracy score) of our previous results. We call this technique top-$k$ analysis, and it describes the configurations with the empirical probability of a parameter to be enabled among the best $k$ configurations.
The score values are defined as the minimum among the set.
The main idea is to discover patterns on the composition of best performing configurations.
As we double $k$ at each row, then $k$ and 2$k$ share $k$ configurations which produces a smoothly convergence to 0.5 for each probability.
At the best of our knowledge, this kind of analysis has never been used in the literature.

\begin{table}[t!]
\caption{Analysis of the $k$ best configurations for the INEGI benchmark in both training and test datasets.}
\label{tab/inegi-topk}
\begin{minipage}{\textwidth}
\resizebox{\textwidth}{!}{
\begin{tabular}{|c|cc|ccc|ccc ccc|ccc ccc|c|}
\hline
k &  accuracy & macro-F1 & tfidf &  del-sw &  lem &  stem &  del-d1 &  del-d2 &  del-punc &  del-diac & del-ent &  emo &  num & url &  usr &   lc &  neg \\
\hline
1   & 0.6065 & 0.4524 & 1.00 &  0.00 & 0.00 & 1.00 & 0.00 & 0.00 & 0.00 &   0.00 &1.00 &   1.00 &   0.00 &   0.00 &   1.00 & 0.00 & 1.00 \\
2   & 0.6065 & 0.4524 & 1.00 &  0.00 & 0.00 & 1.00 & 0.00 & 0.00 & 0.50 &   0.00 &1.00 &   1.00 &   0.00 &   0.00 &   1.00 & 0.00 & 0.50 \\
4   & 0.6065 & 0.4524 & 1.00 &  0.00 & 0.00 & 1.00 & 0.00 & 0.00 & 0.50 &   0.00 &1.00 &   1.00 &   0.00 &   0.00 &   1.00 & 0.00 & 0.50 \\
8   & 0.6059 & 0.4511 & 1.00 &  0.00 & 0.00 & 1.00 & 0.00 & 0.00 & 0.50 &   0.00 &1.00 &   1.00 &   0.50 &   0.00 &   1.00 & 0.00 & 0.50 \\
16  & 0.6058 & 0.4568 & 1.00 &  0.19 & 0.00 & 1.00 & 0.19 & 0.00 & 0.44 &   0.13 &0.81 &   1.00 &   0.38 &   0.19 &   1.00 & 0.19 & 0.5625 \\
32  & 0.6052 & 0.4507 & 1.00 &  0.31 & 0.00 & 0.69 & 0.25 & 0.00 & 0.47 &   0.19 &0.56 &   1.00 &   0.31 &   0.38 &   1.00 & 0.44 & 0.5312 \\
64  & 0.6047 & 0.4516 & 1.00 &  0.22 & 0.00 & 0.78 & 0.44 & 0.00 & 0.50 &   0.33 &0.66 &   1.00 &   0.38 &   0.19 &   1.00 & 0.53 & 0.5156 \\
128 & 0.6037 & 0.4643 & 1.00 &  0.20 & 0.00 & 0.77 & 0.45 & 0.03 & 0.50 &   0.31 &0.53 &   1.00 &   0.42 &   0.28 &   1.00 & 0.58 & 0.4922 \\
256 & 0.6024 & 0.4489 & 1.00 &  0.14 & 0.00 & 0.77 & 0.36 & 0.09 & 0.50 &   0.40 &0.51 &   1.00 &   0.44 &   0.43 &   1.00 & 0.62 & 0.5078 \\
512 & 0.6008 & 0.4315 & 1.00 &  0.17 & 0.00 & 0.73 & 0.42 & 0.17 & 0.50 &   0.43 &0.41 &   1.00 &   0.41 &   0.48 &   0.99 & 0.62 & 0.5098 \\
\hline
\multicolumn{18}{c}{~} \\
\multicolumn{18}{c}{a) Performance on the training dataset (5-folds)} \\
\end{tabular}
}\vspace{0.25cm}
\end{minipage}
\begin{minipage}{\textwidth}
\resizebox{\textwidth}{!}{
\begin{tabular}{|c|cc|ccc|ccc ccc|ccc ccc|c|}
\hline
k &  accuracy &  macro-F1 &  tfidf &  del-sw &  lem &  stem &  del-d1 &  del-d2 &  del-punc &  del-diac & del-ent &  emo &  num & url &  usr &   lc &  neg \\
\hline
1   &    0.6148 &   0.4442 &   1.00 &    0.00 & 0.00 &  0.00 &      0.00 &      0.00 &      0.00 &      1.00 & 0.00 & 1.00 & 0.00 & 0.00 & 1.00 & 1.00 & 1.00 \\
2   &    0.6148 &   0.4442 &   1.00 &    0.00 & 0.00 &  0.00 &      0.00 &      0.00 &      0.50 &      1.00 & 0.00 & 1.00 & 0.00 & 0.00 & 1.00 & 1.00 & 1.00 \\
4   &    0.6136 &   0.4405 &   1.00 &    0.00 & 0.00 &  0.00 &      0.00 &      0.00 &      0.50 &      1.00 & 0.00 & 1.00 & 0.50 & 0.00 & 1.00 & 1.00 & 1.00 \\
8   &    0.6135 &   0.4545 &   1.00 &    0.00 & 0.00 &  0.25 &      0.00 &      0.00 &      0.62 &      0.75 & 0.00 & 1.00 & 0.50 & 0.00 & 1.00 & 1.00 & 0.88 \\
16  &    0.6134 &   0.4546 &   1.00 &    0.00 & 0.00 &  0.38 &      0.00 &      0.00 &      0.50 &      0.88 & 0.00 & 1.00 & 0.62 & 0.38 & 1.00 & 1.00 & 0.81 \\
32  &    0.6130 &   0.4528 &   1.00 &    0.00 & 0.00 &  0.50 &      0.06 &      0.00 &      0.50 &      0.94 & 0.00 & 1.00 & 0.44 & 0.44 & 1.00 & 1.00 & 0.62 \\
64  &    0.6119 &   0.4403 &   1.00 &    0.12 & 0.00 &  0.44 &      0.19 &      0.00 &      0.50 &      0.72 & 0.00 & 1.00 & 0.44 & 0.41 & 1.00 & 1.00 & 0.62 \\
128 &    0.6112 &   0.4547 &   1.00 &    0.30 & 0.00 &  0.48 &      0.27 &      0.00 &      0.50 &      0.61 & 0.00 & 1.00 & 0.50 & 0.52 & 1.00 & 0.98 & 0.61 \\
256 &    0.6099 &   0.4379 &   1.00 &    0.35 & 0.00 &  0.46 &      0.37 &      0.00 &      0.50 &      0.50 & 0.05 & 1.00 & 0.46 & 0.48 & 1.00 & 0.92 & 0.53 \\
512 &    0.6083 &   0.4479 &   1.00 &    0.27 & 0.00 &  0.52 &      0.27 &      0.05 &      0.50 &      0.51 & 0.13 & 1.00 & 0.45 & 0.48 & 1.00 & 0.75 & 0.55 \\
\hline
\multicolumn{18}{c}{~} \\
\multicolumn{18}{c}{b) Performance on the gold-standard dataset} \\
\end{tabular}
}
\end{minipage}
\end{table}

All tables in this subsection are induced by the accuracy score (i.e., best $k$ as measured with accuracy). Also, we display the macro-F1 score as a secondary measure of performance that can help to describe the behaviour of unbalanced multi-class datasets. We omit to show the tokenizer probabilities in favor of Figures~\ref{fig/inegi/qgrams} and \ref{fig/tass/qgrams}; please remind that almost all top configurations use $q$-grams.

Table~\ref{tab/inegi-topk} shows the composition of INEGI's best configurations in both training and test sets. As previously shown,  almost all best setups enable \textsf{TFIDF}, and properly handle emoticons and users. The parameters {\em del-sw, lem, del-d1, del-d2, num,} and {\em url}, are almost deactivated in both training and test sets. The rest of the parameters ({\em stem, del-diac, del-ent,} and {\em neg}) do not remain between training and test sets. However, the later set of parameters are disabled in the gold-standard best configurations, excepting for {\em neg}. Such fact supports the idea that faster configurations also can produce excellent performances. Please notice that lemmatization ({\em lem}) and stemming ({\em stem}) are also disabled, which are the linguistic operations with higher computational costs in our pipeline of text transformations.

\begin{table}[th!]
\caption{Analysis of the $k$ best configurations (top-$k$) for the TASS'15 benchmark in both training and test datasets.}
\label{tab/tass-topk}
\begin{minipage}{\textwidth}
\resizebox{\textwidth}{!}{
\begin{tabular}{|c|cc|ccc|ccc ccc|ccc ccc|c|}
\hline
k &  accuracy &  macro-F1 &  tfidf &  del-sw &  lem &  stem &  del-d1 &  del-d2 &  del-punc &  del-diac & del-ent &  emo &  num & url &  usr &   lc &  neg \\
\hline
1   &    0.6286 &   0.4951 &  1.00 &    1.00 & 0.00 &  0.00 &      1.00 &      0.00 &      1.00 &      1.00 & 0.00 & 0.00 & 1.00 & 1.00 & 1.00 & 1.00 & 1.00 \\
2   &    0.6286 &   0.4951 &  1.00 &    1.00 & 0.00 &  0.00 &      1.00 &      0.00 &      0.50 &      1.00 & 0.00 & 0.00 & 1.00 & 1.00 & 1.00 & 1.00 & 1.00 \\
4   &    0.6281 &   0.4947 &  1.00 &    0.75 & 0.00 &  0.00 &      0.75 &      0.00 &      0.50 &      1.00 & 0.00 & 0.50 & 0.75 & 1.00 & 1.00 & 0.75 & 1.00 \\
8   &    0.6279 &   0.4895 &  1.00 &    0.50 & 0.00 &  0.00 &      0.50 &      0.00 &      0.50 &      1.00 & 0.00 & 0.50 & 0.50 & 1.00 & 1.00 & 0.50 & 1.00 \\
16  &    0.6270 &   0.4864 &  1.00 &    0.38 & 0.00 &  0.00 &      0.38 &      0.06 &      0.44 &      0.88 & 0.00 & 0.50 & 0.50 & 0.88 & 1.00 & 0.50 & 1.00 \\
32  &    0.6265 &   0.4884 &  1.00 &    0.25 & 0.00 &  0.00 &      0.34 &      0.06 &      0.47 &      0.69 & 0.00 & 0.38 & 0.59 & 0.62 & 1.00 & 0.62 & 1.00 \\
64  &    0.6258 &   0.4852 &  1.00 &    0.20 & 0.00 &  0.00 &      0.42 &      0.22 &      0.50 &      0.62 & 0.00 & 0.48 & 0.56 & 0.48 & 0.94 & 0.61 & 0.88 \\
128 &    0.6254 &   0.4862 &  1.00 &    0.20 & 0.00 &  0.00 &      0.46 &      0.27 &      0.48 &      0.67 & 0.00 & 0.56 & 0.59 & 0.38 & 0.81 & 0.68 & 0.77 \\
256 &    0.6247 &   0.4846 &  1.00 &    0.21 & 0.00 &  0.12 &      0.38 &      0.32 &      0.50 &      0.66 & 0.02 & 0.47 & 0.60 & 0.42 & 0.77 & 0.73 & 0.69 \\
512 &    0.6240 &   0.4848 &  1.00 &    0.14 & 0.00 &  0.24 &      0.38 &      0.36 &      0.50 &      0.65 & 0.02 & 0.47 & 0.61 & 0.42 & 0.77 & 0.67 & 0.63 \\
\hline
\multicolumn{18}{c}{~} \\
\multicolumn{18}{c}{a) Performance in the training dataset (5-folds)} \\
\end{tabular}
}
\end{minipage}\vspace{0.25cm}
\begin{minipage}{\textwidth}
\resizebox{\textwidth}{!}{
\begin{tabular}{|c|cc|ccc|ccc ccc|ccc ccc|c|}
\hline
k &  accuracy &  macro-F1 &  tfidf &  del-sw &  lem &  stem &  del-d1 &  del-d2 &  del-punc &  del-diac & del-ent &  emo &  num & url &  usr &   lc &  neg \\
\hline
1   &    0.6330 &   0.5101 &  0.00 &    1.00 & 0.00 &  0.00 &      1.00 &      0.00 &      0.00 &      0.00 & 0.00 & 0.00 & 0.00 & 1.00 & 1.00 & 1.00 & 1.00 \\
2   &    0.6330 &   0.5101 &  0.00 &    1.00 & 0.00 &  0.00 &      1.00 &      0.00 &      0.50 &      0.00 & 0.00 & 0.00 & 0.00 & 1.00 & 1.00 & 1.00 & 1.00 \\
4   &    0.6326 &   0.5099 &  0.00 &    1.00 & 0.00 &  0.00 &      1.00 &      0.00 &      0.50 &      0.00 & 0.00 & 0.50 & 0.00 & 1.00 & 1.00 & 1.00 & 1.00 \\
8   &    0.6317 &   0.5104 &  0.00 &    1.00 & 0.00 &  0.00 &      1.00 &      0.00 &      0.50 &      0.00 & 0.00 & 0.25 & 0.00 & 0.50 & 1.00 & 1.00 & 0.75 \\
16  &    0.6315 &   0.5082 &  0.00 &    1.00 & 0.00 &  0.00 &      1.00 &      0.00 &      0.50 &      0.38 & 0.00 & 0.38 & 0.00 & 0.38 & 1.00 & 1.00 & 0.88 \\
32  &    0.6315 &   0.5069 &  0.00 &    0.69 & 0.00 &  0.12 &      0.78 &      0.16 &      0.47 &      0.38 & 0.00 & 0.56 & 0.00 & 0.62 & 0.69 & 1.00 & 0.81 \\
64  &    0.6311 &   0.5071 &  0.00 &    0.62 & 0.00 &  0.12 &      0.83 &      0.19 &      0.48 &      0.39 & 0.00 & 0.66 & 0.12 & 0.56 & 0.62 & 1.00 & 0.81 \\
128 &    0.6302 &   0.5061 &  0.00 &    0.56 & 0.00 &  0.25 &      0.80 &      0.25 &      0.48 &      0.43 & 0.00 & 0.72 & 0.22 & 0.62 & 0.56 & 1.00 & 0.77 \\
256 &    0.6296 &   0.5054 &  0.00 &    0.38 & 0.00 &  0.27 &      0.69 &      0.34 &      0.50 &      0.47 & 0.00 & 0.73 & 0.23 & 0.62 & 0.38 & 0.94 & 0.65 \\
512 &    0.6286 &   0.5048 &  0.06 &    0.39 & 0.00 &  0.34 &      0.68 &      0.42 &      0.50 &      0.46 & 0.02 & 0.75 & 0.38 & 0.57 & 0.36 & 0.80 & 0.66 \\
\hline
\multicolumn{18}{c}{~} \\
\multicolumn{18}{c}{b) Performance in the gold-standard dataset} \\
\end{tabular}
}
\end{minipage}
\end{table}

Table~\ref{tab/tass-topk} shows the top-$k$ analysis for TASS'15. Again, \textsf{TFIDF} is a common ingredient of the majority of the better configurations in the training set; however, the best ones deactivate this parameter to use only the frequency of the term; reflected in a minimum improvement.
The transformations that remain active in both training and set are {\em del-sw, del-d1, url, usr, lc,} and {\em neg}. The deactivated ones in both sets are {\em lem, stem, del-d2, and del-ent,} and {\em emo}.
The rest of the parameters that change between training and test sets are {\em tfidf, del-diac,} and {\em num}.
Note that as $k$ grows, {\em del-punc} and {\em emo}, are close to be random choices.
It is counterintuitive to see the {\em emo} parameter outside the top-$k$ items, the same happens for the {\em del-ent} parameter. The {\em emo} parameter is used to map emoticons and emojis to sentiments, and {\em del-ent} is an heuristic designed to generalize the sentiment expression in the text (see Table~\ref{tab/parameters}).
This behaviour remember us that, in the end, everything depends on the particular distribution of the dataset.
In general, it is clear that there is no a rule-of-thumb to compute the best configuration. Therefore, a probabilistic approach, as it is the output of top-$k$ analysis, is useful to reduce the cost of the exploration of the configuration space.

\subsection{Improving the Performance with Combination of Tokenizers}

In previous experiments, we performed an exhaustive evaluation of the configuration space; then, to improve over our results we need to modify the configuration space. Instead of adding more complex text transformations, we decide to use more than one tokenizer per configuration. More detailed, there exists 127 possible combinations of tokenizers, that is, the powerset of $$\{2\text{-words}, 1\text{-words}, 3\text{-grams}, 4\text{-grams}, 5\text{-grams}, 6\text{-grams}, 7\text{-grams}\}\text{,}$$ minus the empty set.
For this experiment, we only applied the expansion of tokenizers to the best configurations found in the previous experiments, since performing an exhaustive analysis of the new configuration space becomes unfeasible. The hypothesis is that the previous best configurations will be also compose some of the best configurations in the new space, this is a fair assumption that never get worst under an exhaustive analysis.

\begin{figure*}[ht!]
\subfigure[INEGI] {
    \includegraphics[width=0.5\textwidth]{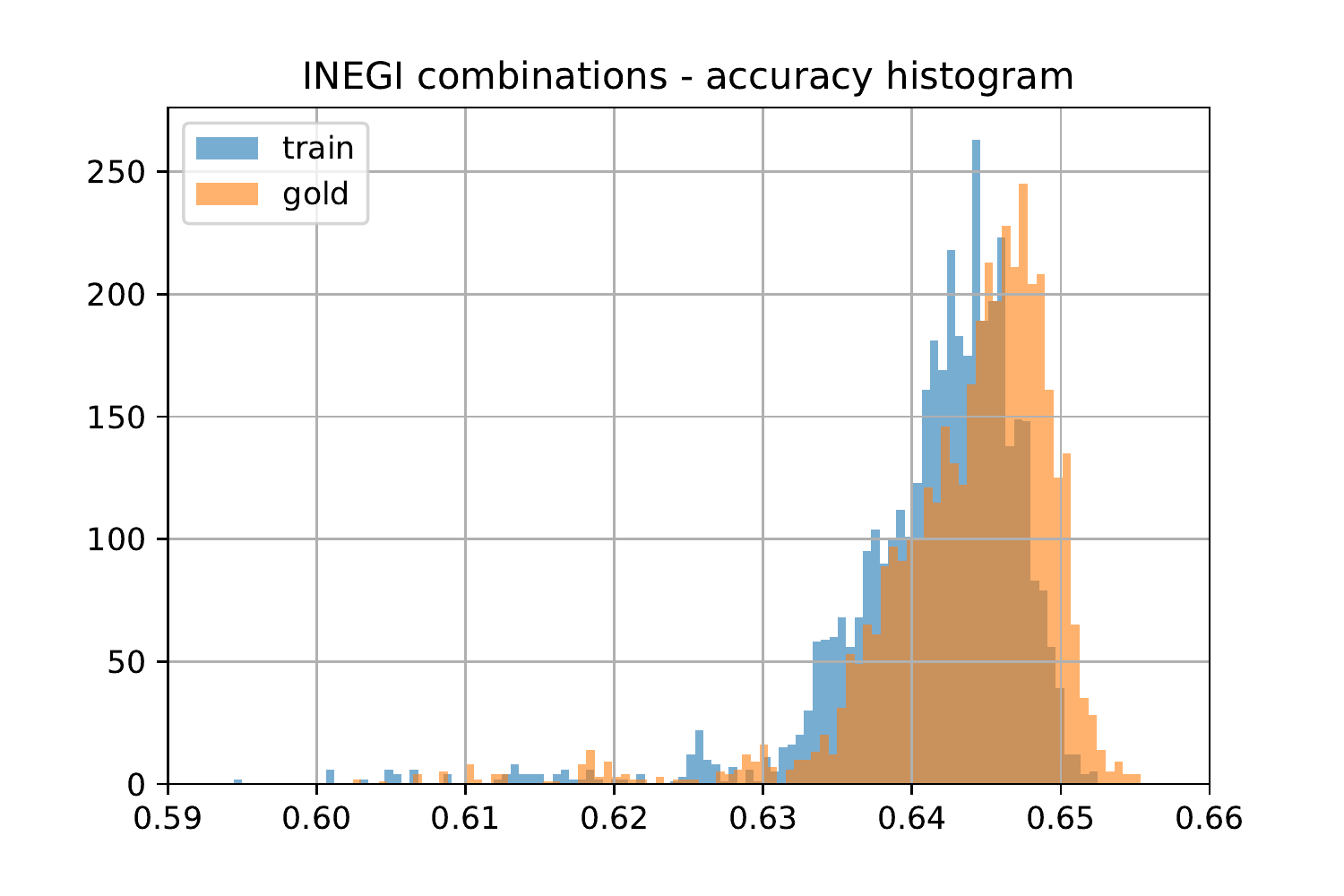}
    \label{fig/inegi/comb}
}
\subfigure[TASS'15] {
    \includegraphics[width=0.5\textwidth]{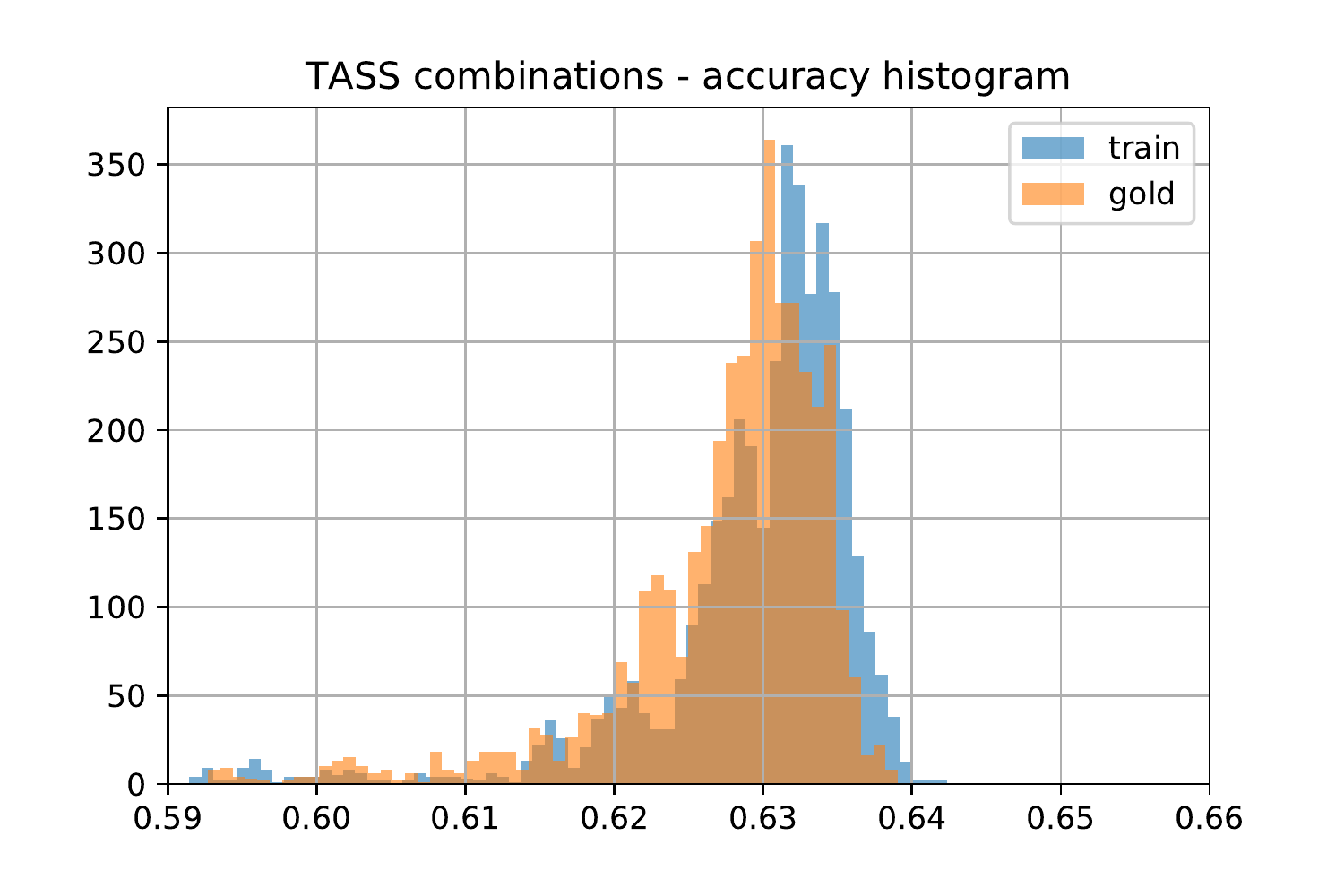}
    \label{fig/tass/comb}
}

\caption{Accuracy's histogram for combination of tokenizers.}
\label{fig/comb/gold}
\end{figure*}

Figure~\ref{fig/inegi/comb} shows the performance of 4064 configurations that correspond to all combinations of tokenizers over the top-$32$ configurations on the training set, see Table~\ref{tab/inegi-topk}. The performance in both training and test sets is pretty similar, and significantly better than that achieved with a single tokenizer (Table~\ref{tab/inegi-topk}).
In Table~\ref{tab/comb/gold} and Figure~\ref{fig/comb/gold} we can see a significant improvement with respective to single tokenizers. The top-$k$ analysis for the test set is listed in Table~\ref{tab/comb/gold}. In this table we focus on describe the composition of the tokenizers, instead of the text transformations. The analysis shows that $1$-words, $2$-words, $3$-grams, $4$-grams, and $7$-grams are commonly present on the best configurations.

We found that TASS'15 also improves its performance under the combination of tokenizers, as Figure~\ref{fig/tass/comb} illustrates. In this case, the performance in the gold standard does not surpasses the performance on the training set, as is the case of INEGI, but it is pretty close. Table~\ref{tab/comb/gold} shows the composition of the configurations, here we can observe that best performances use $1$-words, and $3$-grams, $4$-grams and $6$-grams. It is interesting to note that $2$-words are not used for the top-8 configurations, in contrast to the best configurations for INEGI.

As mentioned, any datasets will need to adjust the configuration and search for the best combination in the training set, and then, apply to their particular gold-standard. This is a costly procedure, but it is possible to reduce the search space to a sample lead by the probability models of the top-$k$ analysis.
The presented top-$k$ analysis are particularly useful for sentiment analysis in Spanish, other languages may present different models but they are beyond the scope of this manuscript.

\begin{table*}[t]
\caption{Analysis of the top-$k$ combinations of tokenizers for both INEGI and TASS'15 benchmarks. We consider both $n$-words and $q$-grams.}
\label{tab/comb/gold}
\begin{minipage}{0.5\textwidth}
\resizebox{\textwidth}{!}{
\begin{tabular}{|c|cc| cc | cc ccc|}
\hline
\multicolumn{10}{|c|}{INEGI} \\ \hline
k &  accuracy &  macro-F1 &  n=2 &  n=1 &  q=3 &  q=4 &  q=5 &  q=6 &  q=7 \\
\hline
1   &    0.6553 &   0.5287 &   1.00 &   1.00 &   1.00 &   1.00 &   0.00 &   0.00 &   1.00 \\
2   &    0.6550 &   0.5270 &   1.00 &   1.00 &   1.00 &   0.50 &   0.00 &   0.00 &   1.00 \\
4   &    0.6549 &   0.5281 &   0.50 &   1.00 &   1.00 &   0.75 &   0.00 &   0.00 &   1.00 \\
8   &    0.6542 &   0.5268 &   0.63 &   1.00 &   1.00 &   0.62 &   0.00 &   0.00 &   1.00 \\
16  &    0.6538 &   0.5263 &   0.75 &   0.94 &   1.00 &   0.75 &   0.00 &   0.00 &   1.00 \\
32  &    0.6527 &   0.5241 &   0.66 &   0.84 &   1.00 &   0.59 &   0.00 &   0.06 &   0.88 \\
64  &    0.6519 &   0.5235 &   0.56 &   0.77 &   1.00 &   0.52 &   0.09 &   0.06 &   0.84 \\
128 &    0.6510 &   0.5258 &   0.65 &   0.61 &   0.99 &   0.55 &   0.18 &   0.25 &   0.78 \\
256 &    0.6502 &   0.5205 &   0.61 &   0.64 &   0.97 &   0.58 &   0.22 &   0.31 &   0.79 \\
512 &    0.6492 &   0.5172 &   0.62 &   0.66 &   0.96 &   0.55 &   0.30 &   0.40 &   0.74 \\
\hline
\end{tabular}}\end{minipage}
\begin{minipage}{0.5\textwidth}\resizebox{\textwidth}{!}{\begin{tabular}{|c|cc| cc | cc ccc|}
\hline
\multicolumn{10}{|c|}{TASS 15} \\ \hline
k &  accuracy &  macro-F1 &  n=2 &  n=1 &  q=3 &  q=4 &  q=5 &  q=6 &  q=7 \\
\hline

1   &    0.6391 &   0.4997 &   0.00 &   1.00 &   1.00 &   1.00 &   0.00 &   1.00 &   0.00 \\
2   &    0.6391 &   0.4995 &   0.00 &   1.00 &   1.00 &   1.00 &   0.00 &   1.00 &   0.00 \\
4   &    0.6391 &   0.4997 &   0.00 &   1.00 &   1.00 &   1.00 &   0.00 &   1.00 &   0.00 \\
8   &    0.6383 &   0.5020 &   0.00 &   1.00 &   1.00 &   0.75 &   0.50 &   0.75 &   0.00 \\
16  &    0.6380 &   0.4966 &   0.25 &   1.00 &   1.00 &   0.75 &   0.38 &   0.63 &   0.13 \\
32  &    0.6373 &   0.4972 &   0.18 &   1.00 &   1.00 &   0.63 &   0.50 &   0.69 &   0.19 \\
64  &    0.6363 &   0.4940 &   0.30 &   1.00 &   0.94 &   0.75 &   0.55 &   0.58 &   0.17 \\
128 &    0.6356 &   0.4937 &   0.32 &   0.97 &   0.94 &   0.77 &   0.53 &   0.66 &   0.26 \\
256 &    0.6347 &   0.4927 &   0.39 &   0.96 &   0.88 &   0.81 &   0.56 &   0.66 &   0.35 \\
512 &    0.6338 &   0.4954 &   0.42 &   0.92 &   0.86 &   0.73 &   0.57 &   0.56 &   0.39 \\
\bottomrule
\end{tabular}
}
\end{minipage}
\end{table*}


\begin{table*}[t!]
\caption{Top-$k$ analysis of a configuration handcrafted to reduce the computational cost.}
\label{tab/simple-comb}
\begin{minipage}{0.5\textwidth}
\resizebox{\textwidth}{!}{
\begin{tabular}{|c|cc| cc | cc ccc|}
\hline
\multicolumn{10}{|c|}{INEGI} \\ \hline
k &  accuracy &  macro-F1 &  n=2 &  n=1 &  q=3 &  q=4 &  q=5 &  q=6 &  q=7 \\
\hline
1   &    0.6546 &   0.5279 &   1.00 &   1.00 &   1.00 &   1.00 &   0.00 &   0.00 &   1.00 \\
2   &    0.6538 &   0.5268 &   1.00 &   1.00 &   1.00 &   0.50 &   0.00 &   0.00 &   1.00 \\
4   &    0.6525 &   0.5266 &   0.50 &   1.00 &   1.00 &   0.50 &   0.00 &   0.00 &   1.00 \\
8   &    0.6519 &   0.5257 &   0.63 &   0.75 &   1.00 &   0.50 &   0.25 &   0.00 &   1.00 \\
16  &    0.6513 &   0.5237 &   0.69 &   0.75 &   0.94 &   0.56 &   0.25 &   0.31 &   0.88 \\
32  &    0.6503 &   0.5270 &   0.59 &   0.66 &   0.94 &   0.56 &   0.31 &   0.41 &   0.75 \\
64  &    0.6478 &   0.5225 &   0.55 &   0.61 &   0.78 &   0.61 &   0.47 &   0.59 &   0.67 \\
96  &    0.6435 &   0.5250 &   0.55 &   0.55 &   0.61 &   0.60 &   0.54 &   0.54 &   0.57 \\
120 &    0.6412 &   0.5128 &   0.54 &   0.54 &   0.55 &   0.54 &   0.55 &   0.54 &   0.55 \\
127 &    0.5736 &   0.3946 &   0.50 &   0.50 &   0.50 &   0.50 &   0.50 &   0.50 &   0.50 \\
\hline
\end{tabular}}\end{minipage}
\begin{minipage}{0.5\textwidth}\resizebox{\textwidth}{!}{\begin{tabular}{|c|cc| cc | cc ccc|}
\hline
\multicolumn{10}{|c|}{TASS 15 } \\ \hline
k &  accuracy &  macro-F1 &  n=2 &  n=1 &  q=3 &  q=4 &  q=5 &  q=6 &  q=7 \\
\hline
1   &    0.6364 &   0.4971 &   0.00 &   1.00 &   1.00 &   1.00 &   0.00 &   1.00 &   0.00 \\
2   &    0.6357 &   0.4943 &   0.00 &   1.00 &   1.00 &   1.00 &   0.50 &   0.50 &   0.50 \\
4   &    0.6350 &   0.4920 &   0.00 &   1.00 &   1.00 &   0.75 &   0.25 &   0.50 &   0.50 \\
8   &    0.6343 &   0.4948 &   0.25 &   1.00 &   1.00 &   0.75 &   0.50 &   0.63 &   0.25 \\
16  &    0.6336 &   0.4943 &   0.38 &   0.94 &   0.94 &   0.69 &   0.63 &   0.63 &   0.25 \\
32  &    0.6319 &   0.4890 &   0.44 &   0.84 &   0.81 &   0.78 &   0.59 &   0.53 &   0.44 \\
64  &    0.6296 &   0.4842 &   0.47 &   0.69 &   0.73 &   0.70 &   0.59 &   0.55 &   0.47 \\
96  &    0.6252 &   0.4895 &   0.49 &   0.58 &   0.60 &   0.63 &   0.57 &   0.53 &   0.50 \\
120 &    0.6207 &   0.4748 &   0.50 &   0.54 &   0.55 &   0.56 &   0.56 &   0.54 &   0.51 \\
127 &    0.5471 &   0.4154 &   0.50 &   0.50 &   0.50 &   0.50 &   0.50 &   0.50 &   0.50 \\
\hline
\end{tabular}
}
\end{minipage}
\end{table*}

It is worth to mention that the best performance is high dependent of the particular dataset; however, based on Tables~\ref{tab/inegi-topk} and \ref{tab/tass-topk}, it is interesting to note that simpler configurations are among the best performing ones when $q$-grams are used as tokenizers.
This allows to create a model that reduces the computational cost and even improves the performance of the top-1 of both, INEGI and TASS'15, datasets with a single tokenizer.
We create a configuration created by activate {\em tfidf, emo, num, usr,} and {\em lc}; and deactivate {\em del-sw, lem, stem, del-d1, del-d2, del-punc, del-diac, del-ent,} and {\em neg}. All the activated parameters are relatively simple to implement, even without external libraries. Note that leaving out stemming and lemmatization dramatically reduces many times evaluation time.

Table~\ref{tab/simple-comb} shows the performance on the test set. The best configuration based on single tokenizer is $0.6148$ and $0.6330$ for INEGI and TASS'15, respectively; the best performance for combination of tokenizers is $0.6553$ and $0.6391$, in the same order. For our handcrafted configuration we reach an accuracy of $0.6546$ for INEGI, and $0.6364$ for TASS'15. This is very competitive if we take into account that the model selection is reduced to evaluate 127 configurations, and also, each evaluation is pretty fast among other alternatives.

The performances of this simple configurations are pretty close to the best possible ones with our scheme, that is, the gold-standard performance shown in Tables~\ref{tab/inegi-topk} and \ref{tab/tass-topk} while it can be easily implemented and optimized.

\subsection{Performance Comparison on the TASS'15 Challenge}
In the end, a sentiment classifier is a tool that helps to discover the opinion of a crowd of people, the effectiveness is crucial.
So, there exists many researchers interested in the field, and for instance, TASS'15 (\cite{tass2015}) is a forum that gathers many practitioners and researchers for the Spanish version of the problem. As described in \S\ref{sec:Related_Work}, the problem is commonly tackled with the use of affective dictionaries, distant supervision methods to increase the knowledge database, word-embedding techniques, complex linguistic tools like lemmatizers, deep learning based classifiers, among other sophisticated techniques.
Beyond the use of the SVM, there is no complex procedure that limits the adoption of our approach only to expert users.

\begin{figure}[t!]
\centerline{\includegraphics[width=0.6\textwidth]{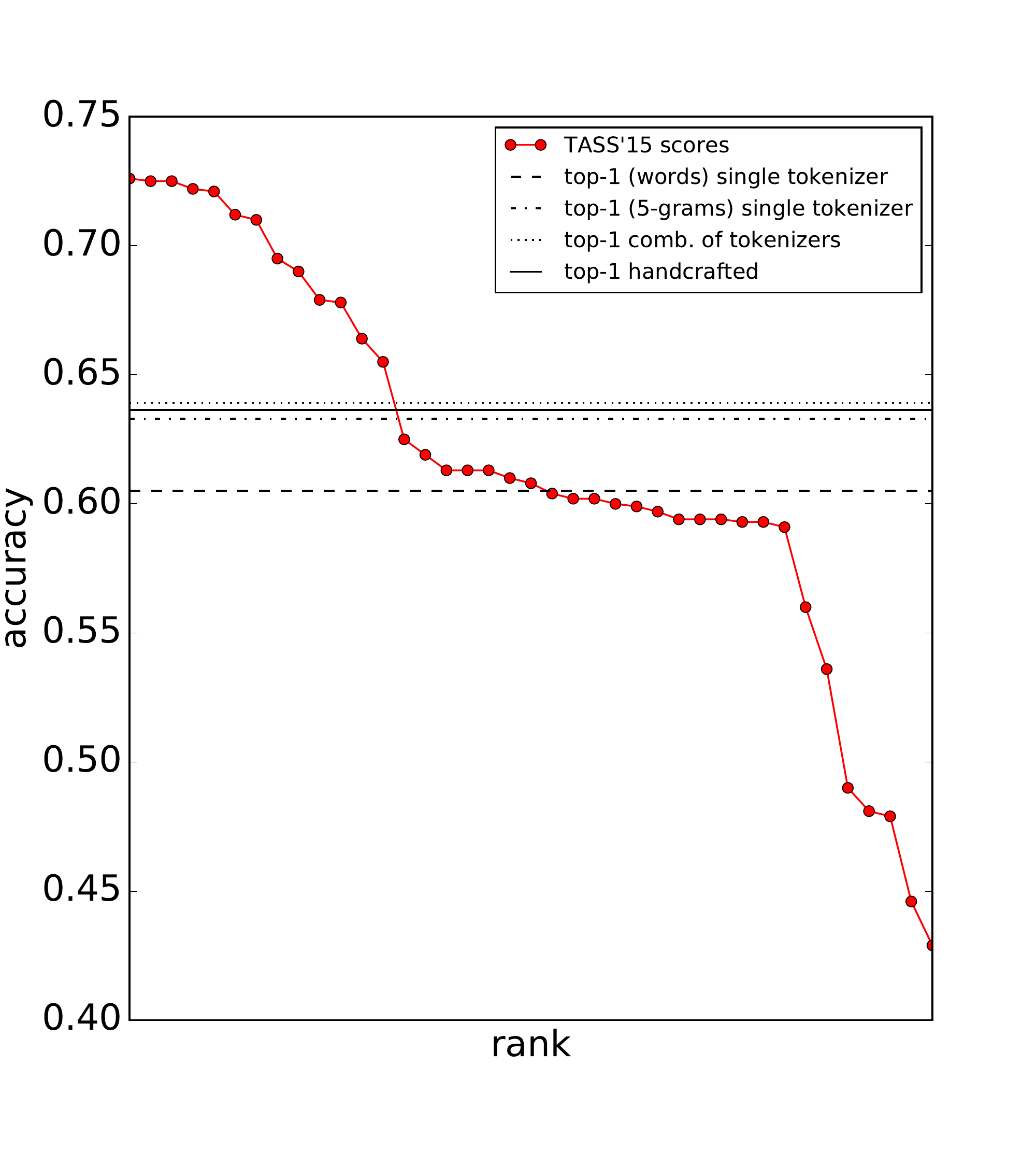}}
\caption{Comparison of our sentiment classifiers with the final scores of TASS'15.}
\label{fig/official-tass2015}
\end{figure}

However, the question is, how good our approach is as compared with both the state-of-the-art and the state-of-the-technique? We use the TASS'15 benchmark to answer this question.
Section~\ref{sec:Related_Work} reviews several of the best papers in the workshop. Figure~\ref{fig/official-tass2015} shows the official scores of TASS'15 participants, the best scores achieve 0.72 and the worst ones are below 0.43. The gross of the participants are between $0.59$ and $0.61$; there lies the best sentiment classifier based on $n$-words ($0.6051$).
The best configuration that uses $q$-grams, as a single tokenizer, surpasses that range, i.e., $0.6330$. The classifiers based on the combination of tokenizers produce a slightly better performances, and our configuration handcrafted for speed is not too distant from these performances, as figure shows.

The magnitude of the improvement is tightly linked to the dataset; for instance, as compared with the best $n$-words sentiment classifier, the performance of INEGI is improved in $11.17\%$ after applying the combination of tokenizers. In the case of TASS'15, the improvement is of $5.62\%$, smaller but significant in any case. It is important to take into account this effect in the design of new sentiment classifiers.

\section{Discussion}
\label{sec:discussion}
In this study, we covered many traditional techniques used to prepare text representations for sentiment analysis. The majority of them are too simple to be aware of their complexities. However, it is important to know its contribution to the solution of the task being tackled, as we showed, sometimes applying some technique is counterproductive.
Therefore, the transformation pipeline should be carefully prepared.
Other techniques, like lemmatization and stemming, are too complex to be implemented each time they are needed; therefore, a mature implementation should be used. However, as our experimental results support, for the sentiment analysis task in Spanish, there is no need to use these complex linguistic techniques if our approach, based on the combination of tokenizers, is used.

More detailed, a lemmatizer is tightly linked to the language being processed, we use Freeling by~\cite{freeling} for Spanish, and it is designed to work on mostly well-written text. The stemming procedure is another sophisticated tool, in our case, we used the Snowball for Spanish, available in NLTK package by~\cite{NLTK2009}. Since it is based mostly on the removal of suffixes, then it is more robust to errors than a lemmatizer. Both techniques are computationally expensive, and both are not used by best-performing configurations; therefore, they should not be applied when the text is full of errors. This is the case of Twitter, the source of our data.

From the perspective of practitioners, the simpler approach is to find the best tokenizer's combination as applied to a set of simple setups; this gives us $127$ combinations if our $\{\text{2-word, 1-word, 3-gram, 4-gram, 5-gram, 6-gram, 7-gram}\}$ set is used. Supported by the patterns found in our top-$k$ analysis, the combinations should have at least three tokenizers, and $1$-words and $3$-grams can always be selected. So, if the complexity of the model selection is an issue, only ${5 \choose 3}+{5\choose 4}+{5\choose 5} = 16$ combinations are needed.

\section{Conclusions}
\label{sec:conclusions}
We were able to improve the performance of our sentiment classifiers significantly. Our approach is simple; given a good {\em initial configuration}, we can enhance its performance using a set of tokenizers that include both $n$-words and $q$-grams. We exhaustively prove the superiority of $q$-grams over $n$-words, at least for our case of study (sentiment analysis in the Spanish language). At first glance, large $q$-grams ($q=5, 6,$ or $7$) are quasi-words; however, the $q$-grams are sliding windows over the entire text, meaning that many times they cover the connection between two words or even three words. In relatively large words, the suffixes and prefixes are captured, when $q$ is small, affixes and word's root are also captured. Nonetheless, this process creates many noisy substrings, and that is the reason behind our best configurations almost always use \textsf{TFIDF}, which weights the tokens to reduce this effect. It is necessary to produce a better process to filter out tokens that not contribute beyond creating larger vectors.

However, a na\"ive implementation of the multiple tokenizers will multiply the necessary memory, i.e., actually it increases the memory needs by a factor of $q$ for $q$-grams. This can be a problem on very large collections. Further research is needed to solve this issue.

The {\em initial configuration} can be a little tricky. In this study, we provide several top-$k$ analysis; the tables produced can be seen as probabilistic models to create good performing classifiers. These models should be valid at least for Spanish.
In practice, this means that we need to evaluate the performance of a few dozens of configurations to select the best performing one among them.  In a modern multicore computing architecture, this means a relatively fast procedure.


Finally, we conjectured that our approach would generalizes to different languages because it works using a few language-specific techniques. However, this claim should be supported by experimental evidence. Also, we provide a list of simple rules to find a sentiment classifier based on our findings; nonetheless, the best setup is dependent of the dataset, the classes, and many others task-dependent properties. In this paper, our approach consists in performing an exhaustive evaluation of the parameter's space and then expand the search using a combination of tokenizers. We will require a faster algorithm to find good setups on large configuration's spaces that work on different languages. Finally, we want to make evident that we used SVM  as classifier because of its popularity in the community, this paper mainly focuses on the treatment of the text regardless, so the proper selection and tuning of the classifier is left as future work.

\section*{Acknowledgements}
The authors would like to thank the anonymous reviewers for their valuable comments and suggestions to improve the quality of this manuscript. We want to thank the {\em Instituto Nacional de Estad\'istica y Geograf\'ia} (INEGI) of M\'exico for granting access to the labeled-benchmark of INEGI and its dataset of geolocated tweets, especially to
to Gerardo Leyva, Juan Muñoz, Alfredo Bustos, Silvia Fraustro, and Abel Coronado.
We also would like to thank Julio Villena-Roman for kindly give us access to the gold-standards of TASS'15.

\section*{\refname}
\bibliographystyle{apalike}
\bibliography{bibused}

\end{document}